\documentclass[runningheads]{llncs}

\usepackage{eccv}

\usepackage{eccvabbrv}

\usepackage{graphicx}
\usepackage{booktabs}
\usepackage{comment}
\usepackage{color}
\usepackage{multirow}
\usepackage{listings}
\usepackage{subcaption}
\usepackage{amssymb}
\usepackage{bm}

\usepackage[accsupp]{axessibility}  

\usepackage{hyperref}

\usepackage{orcidlink}

\definecolor{codegreen}{rgb}{0,0.6,0}
\definecolor{codegray}{rgb}{0.5,0.5,0.5}
\definecolor{codepurple}{rgb}{0.58,0,0.82}
\definecolor{backcolour}{rgb}{0.95,0.95,0.92}

\lstdefinestyle{mystyle}{
  backgroundcolor=\color{backcolour}, commentstyle=\color{codegreen},
  keywordstyle=\color{magenta},
  numberstyle=\tiny\color{codegray},
  stringstyle=\color{codepurple},
  basicstyle=\ttfamily\scriptsize,
  breakatwhitespace=false,         
  breaklines=true,                 
  captionpos=b,                    
  keepspaces=true,                 
  numbers=left,                    
  numbersep=5pt,                  
  showspaces=false,                
  showstringspaces=false,
  showtabs=false,                  
  tabsize=2,
}

\begin{document}

\title{
Data Collection-free Masked Video Modeling \\
}


\author{
Yuchi Ishikawa\inst{1,2}\orcidlink{0000-0002-9485-6840} \and
Masayoshi Kondo\inst{1}\orcidlink{0009-0003-6485-2480} \and
Yoshimitsu Aoki\inst{2}\orcidlink{0000-0001-7361-0027}
}

\authorrunning{Y.~Ishikawa et al.}

\institute{
LY Corporation\\
\email{\{yuchi.ishikawa, masayoshi.kondo\}@lycorp.co.jp} \and
Keio University\\
\email{aoki@elec.keio.ac.jp}
}

\maketitle

\begin{abstract}
    Pre-training video transformers generally requires a large amount of data,
    presenting significant challenges in terms of data collection costs
    and concerns related to privacy, licensing, and inherent biases.
    Synthesizing data is one of the promising ways to solve these issues,
    yet pre-training solely on synthetic data has its own challenges.
    In this paper, we introduce an effective self-supervised learning
    framework for videos that leverages readily available and less costly
    static images.
    Specifically, we define the Pseudo Motion Generator (PMG) module
    that recursively applies image transformations to generate pseudo-motion
    videos from images.
    These pseudo-motion videos are then leveraged in masked video modeling.
    Our approach is applicable to synthetic images as well, thus entirely
    freeing video pre-training from data collection costs and other concerns
    in real data. 
    Through experiments in action recognition tasks, we demonstrate that
    this framework allows effective learning of spatio-temporal features through
    pseudo-motion videos,
    significantly improving over existing methods which also use static images
    and partially outperforming those using both real and synthetic videos.
    These results uncover fragments
    of what video transformers learn through masked video modeling.
\end{abstract}

\section{Introduction}

Pre-training video transformers~\cite{arnab2021vivit, bertasius2021space} 
generally requires a large amount of labeled data.
Although self-supervised learning enables pre-training of video transformers
without labels~\cite{tong2022videomae,feichtenhofer2022masked},
it still demands substantial volumes of video data.
This highlights various issues related to real video data including the following:

\vspace{-1mm}
\begin{description}
    
\item{\textbf{High cost of data collection.}}
Video data, compared to audio, text, and images, is massive in size.
Therefore, downloading, storing, and pre-processing videos
is extremely costly.
Furthermore, the following issues related to licenses, privacy, and bias arise during data collection.

\item{\textbf{Copyright and license infringement.}}
Video data may have been collected without permission,
potentially infringing on licenses and copyrights.
For example, some datasets are gathered from video-sharing sites
like YouTube, in which videos are licensed by default with a
Standard YouTube license\footnote{https://www.youtube.com/t/terms},
which prohibits the download of content.

\item{\textbf{Privacy issues.}}
Video data often contains Personally Identifiable Information (PII)
including faces, which raises significant privacy concerns.

\item{\textbf{Bias and ethical issues.}}
Large-scale datasets may unintentionally include biases
leading to ethical issues related to 
nationality, gender, age, and more~\cite{yang2020towards,
buolamwini2018gender,wang2019balanced},
which can impact the fairness and inclusiveness of model outcomes.
Some works have also reported that video recognition models
might have context and object biases,
failing to recognize actions accurately~\cite{li2018resound,
choi2019can,li2019repair}.

\item{\textbf{Data access issues.}}
Possibly due to the above issues,
some datasets like
IG-Curated/Uncurated dataset~\cite{ghadiyaram2019large,feichtenhofer2021large}
and CREATE~\cite{zhang2022create,ma2023order}
are only made available to certain research groups.
This limitation restricts other researchers from replicating
or further developing these works, thereby impeding scientific progress.

\end{description}

\begin{table}[t]
\centering
\footnotesize

\caption{
    \textbf{Comparison of each data source and their issues when conducting pre-training.}
    While real videos enhance model performance, they have concerns related to
    collection cost, privacy and licenses.
    Synthetic videos~\cite{kataoka2022spatiotemporal,kim2022transferable,zhong2023learning} and
    pseudo motions by MoSI~\cite{huang2021self} partially resolve these issues, but they
    rely on the CNN architecture and its inherent inductive bias,
    thus failing to accurately train ViT.
    Note that in VPN, additional real data is required for optimal performance, therefore
    the asterisked issues (\raisebox{1pt}{\footnotesize$\ast$}) are not resolved.
    Our proposed framework is free of these issues
    by generating pseudo-motion videos from synthetic images.
}

\vspace{-3mm}

\begin{tabular}{ccccc}
    & Real Video~\cite{tong2022videomae}
    & VPN~\cite{kataoka2022spatiotemporal}
    & MoSI~\cite{huang2021self}
    & Ours \\ \hline
example
& \raisebox{-4mm}{\includegraphics[width=0.20\textwidth]{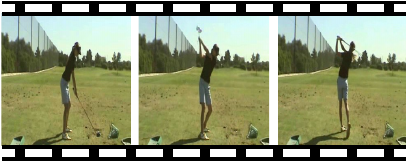}}
& \raisebox{-4mm}{\includegraphics[width=0.20\textwidth]{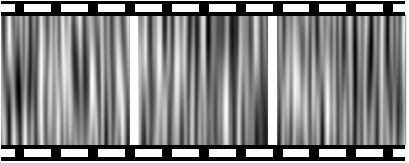}}
& \raisebox{-4mm}{\includegraphics[width=0.20\textwidth]{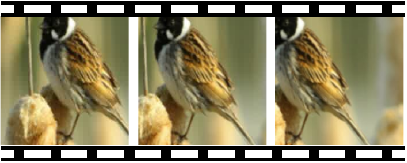}}
& \raisebox{-4mm}{\includegraphics[width=0.20\textwidth]{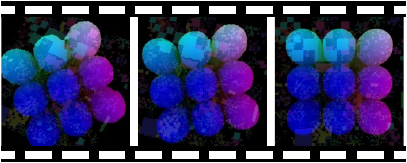}} \\ \hline
Acc. @UCF101           & 96.1        & 89.9                  & 82.8          & 89.4      \\
collection cost        &             & $\checkmark^{\ast}$   & \checkmark    & \checkmark      \\
privacy/license        &             & $\checkmark^{\ast}$   & \checkmark    & \checkmark       \\
training ViT           & \checkmark  &                       &               & \checkmark       \\  \hline

\end{tabular}
\label{table:top}

\vspace{-3mm}

\end{table}

\vspace{-2mm}
In image recognition, to address these concerns
and eliminate the costs associated with data collection,
some researchers have proposed pre-training methods
using synthetic images as an alternative to those using real images.
While some works have synthesized images 
from mathematical formulas~\cite{kataoka2020pre,
takashima2023visual,nakashima2022can},
others have utilized structured noise~\cite{baradad2021learning} or
OpenGL fragment shaders~\cite{baradad2022procedural}.
These methods have achieved comparable results
to pre-training on real image datasets
like ImageNet~\cite{deng2009imagenet} and JFT-300M~\cite{sun2017revisiting},
emphasizing the importance of data diversity.

However,
pre-training using synthetic videos still presents significant challenges.
Few works address this, including the Video Perlin Noise (VPN) dataset~\cite{kataoka2022spatiotemporal}
generated from Perlin Noise~\cite{perlin1985image, perlin2002improving},
and SynAPT~\cite{kim2022transferable,zhong2023learning}.
However,
they still require real video datasets such as Kinetics400~\cite{kay2017kinetics}.
This diverges from our goal of reducing data collection costs
and minimizing issues related to real data.

An alternative approach involves generating pseudo-motions from static images.
Huang et al.~\cite{huang2021self} proposed a self-supervised learning
framework named Unmasked MoSI,
designed to make models learn spatio-temporal features
through the classification of pseudo motions.
This can be promising
because it only requires static images
and can be combined with datasets
with protected
privacy and liberal licenses, like PASS~\cite{asano2021pass}.
However, this method is specialized on CNN architectures
and cannot generalize to transformer-based architectures,
which are the current state-of-the-art models.

In this paper, to mitigate video collection costs
and address concerns regarding privacy, bias, and licenses,
we propose a self-supervised learning framework for video transformers using synthetic images
(Table~\ref{table:top}).
Our framework includes a Pseudo Motion Generator (PMG) module
that recursively applies image transformations to static images,
generating videos with diverse pseudo-motion.
These videos are then used for masked video modeling.
Through experiments, by using videos generated from the PMG module, we examine that video
transformers can learn transferable and robust video features which are not limited to a single domain.
To the best of our knowledge,
we are the first to pre-train video transformers
exclusively using synthetic images.
Our contributions are threefold;

\vspace{-1.5mm}
\begin{enumerate} 
    \item We introduce
    a self-supervised learning framework for videos 
    that uses single images to reduce data collection costs compared to videos.
    Our framework includes a Pseudo Motion Generator (PMG) module,
    which generates a wide variety of pseudo-motion videos.
    These pseudo-motion videos are utilized for self-supervised masked video modeling.
    Notably, PMG can also be used for video augmentation when pre-training with real videos.

    \item We demonstrate that synthetic images can be used for our framework to
    still effectively pre-train video transformers,
    completely eliminating the need for real videos or images.
    This mitigates privacy, bias, and licensing concerns.

    \item Through experiments in action recognition tasks,
    we demonstrate that our proposed framework significantly improves over existing works using static images,
    and also partially surpasses existing pre-training methods
    using both real and synthetic videos.
    These experimental findings reveal pieces of
    what video transformers learn through masked video modeling.
\end{enumerate}

\vspace{-3mm}
\section{Related Work}

\vspace{-1mm}
\noindent\textbf{Self-supervised Learning for Videos.}
Videos require significantly more effort than images and text
for annotation.
Therefore, more interest is invested in self-supervised
learning methods which do not require labeled data.
While earlier works leverage pretext tasks~\cite{yao2020video,wang2020self,jenni2020video,
benaim2020speednet,fernando2017self,xu2019self,ahsan2019video,kim2019self},
recent advancements have introduced
contrastive learning~\cite{feichtenhofer2021large,pan2021videomoco}
and masked video modeling~\cite{bandara2023adamae,wei2022masked,wang2022bevt,
yang2022self,wang2023videomae,li2023unmasked,wang2023masked,sun2023masked},
which offers more robust representation learning without explicit labeling.
Notably, VideoMAE~\cite{feichtenhofer2022masked,tong2022videomae}
has emerged as a leading method due to its simplicity and efficacy,
learning video representations by simply reconstructing masked regions.
Some works, however, point out that VideoMAE predominantly 
learns low-level features
such as shapes.
This tendency may limit its ability to capture
high-level semantic features~\cite{li2023unmasked,shu2022masked}.
Nonetheless, the emphasis on low-level features suggests that
VideoMAE does not specialize in domain-specific features,
leading to its high transferability across various domains.
We aim to capitalize on this characteristic
to train video transformers with static images.

\vspace{1mm}
\noindent\textbf{Large-scale Datasets in Computer Vision. }
Though self-supervised learning eliminates the need for annotation,
it still demands large volumes of data.
The growth of computer vision has relied on massive datasets
like ImageNet~\cite{deng2009imagenet}
and LAION-5B~\cite{schuhmann2022laion}.
However, these resources are fraught with privacy, bias, and licensing issues.
Furthermore, access to datasets like JFT-300M~\cite{sun2017revisiting},
Instagram-3.5B~\cite{mahajan2018exploring},
IG-Curated/Uncurated~\cite{ghadiyaram2019large,feichtenhofer2021large},
and CREATE~\cite{zhang2022create,ma2023order}
is restricted to certain research groups.
These issues underscore the urgent need for accessible data sources
which are free from bias and privacy violation.

Video data exacerbates these challenges with its higher collection costs,
privacy risks and biases~\cite{li2018resound, choi2019can,li2019repair}.
Some popular datasets such as Kinetics400~\cite{kay2017kinetics},
HowTo100M~\cite{miech2019howto100m}, YouTube-8M~\cite{abu2016youtube},
and ActivityNet~\cite{caba2015activitynet},
are collected from YouTube and may encounter copyright and license restrictions.
On the other hand, our self-supervised framework
requires only synthetic images which are free from these challenges.

\vspace{1mm}
\noindent\textbf{Learning from Synthetic Data. }
In response to these challenges,
there is a growing interest in synthetic data,
which bypasses many of the issues existent when using real-world data.
Some research has focused on synthesizing realistic data 
~\cite{wu2018towards,fischer2015flownet,teed2020raft,
hwang2021eldersim,varol2021synthetic,de2016procedural,
mu2020learning},
while others have proposed systematic synthesis of data
from noise~\cite{baradad2021learning} or mathematical formulas
~\cite{kataoka2020pre,nakashima2022can,baradad2022procedural,
takashima2023visual,nakamura2023pre}.
These works have proven that not only realism
but also diversity in synthetic data is crucial for effectively
training models.

Few attempts are made to train action recognition models using synthetic data.
For example, the GATA dataset~\cite{guo2022learning},
collected from a video game,
is proposed for human motion representation learning.
However, this dataset is not allowed for commercial use,
and the rights of game companies have not been considered.
Another example is the Video Perlin Noise (VPN)
dataset~\cite{kataoka2022spatiotemporal},
which is generated from Perlin Noise~\cite{perlin1985image,perlin2002improving}.
This dataset is proposed to initialize model weights before pre-training.
Zhong et al.~\cite{zhong2023learning} propose a pre-training method
with both No-Human Kinetics (NH-Kinetics) and SynAPT~\cite{kim2022transferable}.
While these approaches contribute to model performance,
they still require pre-training on real videos.
Additionally,
ElderSim~\cite{hwang2021eldersim},
PHAV~\cite{de2016procedural},
and SURREAL~\cite{varol2021synthetic},
which are included in SynAPT,
are not allowed for commercial use.
As an alternative,
Huang et al.~\cite{huang2021self} have proposed MoSI,
which pre-trains models with pseudo-motion videos generated from static images.
In terms of collection cost, requiring only static images for pre-training is favorable.
However, because MoSI's synthesized videos lack diversity,
they fail to pre-train video
transformers (See~\cref{tab:comparison_with_sota}).

Overall, existing works have shown the capability
to pre-train video recognition models using synthetic or pseudo-motion videos.
However, they either specialize on CNN architectures or still require the
use of real video data.
In contrast, our method generates a diversity of pseudo-motion videos
from synthetic images,
which can effectively pre-train video transformers.
Moreover, our approach is completely agnostic of the issues
associated with video data collection, privacy, and bias.

\vspace{-2mm}
\section{Proposed Method}

\begin{figure}[t]
    \centering
    \footnotesize
    \includegraphics[width=0.9\textwidth]{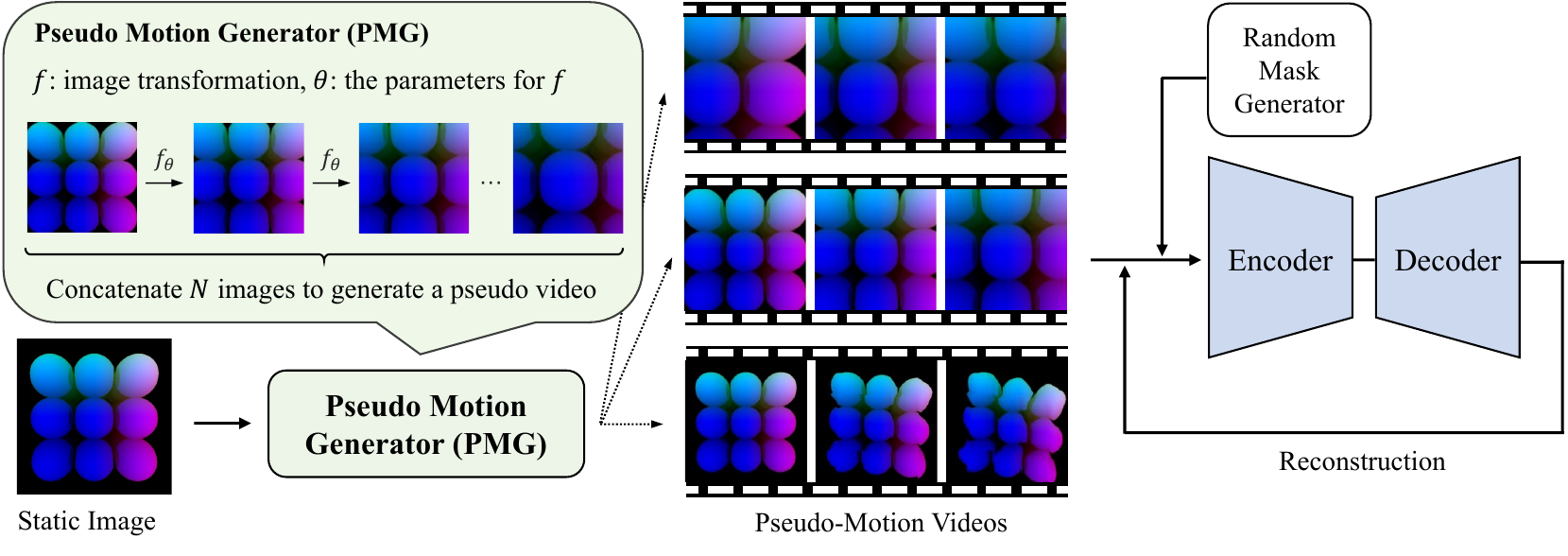}
    \vspace{-2mm}
    \caption{\textbf{Overview of our proposed framework.}}
    \label{fig:proposed_pm_mae}

    \vspace{-4mm}

\end{figure}

\vspace{-2mm}
\subsection{Overview of Our Self-supervised Framework}
\label{sec:overview}

To reduce the collection cost of video data,
we propose 
a self-supervised framework using pseudo-motion videos generated from static images.
Figure~\ref{fig:proposed_pm_mae} shows the overview of our framework.
We first generate pseudo-motion videos from static images by Pseudo Motion Generator (PMG).
Then, we utilize these videos to train VideoMAE~\cite{tong2022videomae, feichtenhofer2022masked}.
VideoMAE is a powerful self-supervised learning framework
and can learn spatio-temporal features effectively
by reconstructing masked video regions from their complementaries.
Some works point out that VideoMAE has a tendency
to learn low-level features such as edges, 
thus failing to achieve high-level alignment~\cite{li2023unmasked, shu2022masked}.
Conversely, VideoMAE does not obtain domain-specific features,
leading to high transferability.
We focus on and leverage this characteristic to train video transformers with pseudo-motion videos.

\vspace{-1mm}
\subsection{Pseudo Motion Generator (PMG)}
\label{sec:pmg}

As mentioned, VideoMAE learns low-level features such as edges in a video.
Especially, we hypothesize that it focuses on the correspondence of patches 
between frames.
Therefore, we assume that to train VideoMAE effectively,
patches between frames in videos should be trackable.
To generate such a pseudo-motion video $V \in \mathbb{R}^{C \times T \times H \times W}$
\footnote{
$T$ is the number of frames in a video.
$H$ and $W$ are the width and the height of each video frame.
$C$ is the number of channels.
},
we propose a simple module, namely Pseudo Motion Generator (PMG).
The algorithm of PMG is as follows:
First, PMG randomly selects an image transformation $f$
from a predefined set $\phi$
and determines its intensity parameter $\theta$.
Then, PMG takes as input a static image $I_1 \in \mathbb{R}^{C \times H \times W}$
and recursively applies image transformation to $I_1$.

\begin{equation}
I_{i+1} = f_{\theta}(I_i) \quad \text{for} \quad i=1,...,T-1
\end{equation}

\noindent Finally, by concatenating the images from $I_1$ to $I_{T}$
in the temporal dimension, a pseudo-motion video $V$ is generated.

\begin{equation}
V = [I_1; I_2; ... ; I_T]
\end{equation}

\noindent For clarity, we also provide pseudo-code for PMG in the supplementary material.

As candidates for image transformation,
we consider the following 8 image transformations.
(See~\cref{sec:effect_of_image_augmentation} for the effect of each transformation)

\vspace{-2.5mm}
\begin{enumerate}
    \item \textbf{Identity: } Return an input image as is. We regard this as a baseline.
    \item \textbf{Sliding Window: }
    Cut a window from a static image and move it randomly.
    Note that this is similar to Unmasked MoSI~\cite{huang2021self},
    but our method does not limit the window's movement to only four directions.

    \item \textbf{Zoom-in/out:}
    Cut a window from an input image and enlarge or reduce the size of the window.

    \item \textbf{Fade-in/out: }
    An input image gradually becomes visible or invisible.

    \item \textbf{Affine Transformation}

    \item \textbf{Perspective Transformation}

    \item \textbf{Color Jitter}: Randomly change the brightness, contrast, saturation, and hue of an input image.

    \item \textbf{CutMix}: Generate an image using CutMix~\cite{yun2019cutmix} from two images and move a small area of the image in the manner of Sliding Window.
\end{enumerate}

\noindent 
From these candidates, through experimentation,
we identify the optimal set of image transformations $\phi$
(See~\cref{sec:combination_of_image_aug}).
Furthermore, to prevent overfitting to specific types of pseudo-motion videos,
we apply mixup~\cite{zhang2017mixup} to each frame of the generated pseudo-motion videos.
This approach significantly enhances the diversity
in motion and appearance of the pseudo-motion videos,
facilitating more efficient learning by VideoMAE.
In the supplementary material, we describe the parameters for each image augmentation.




\begin{figure}[t]
    \centering
    \footnotesize
    \includegraphics[width=0.90\textwidth]{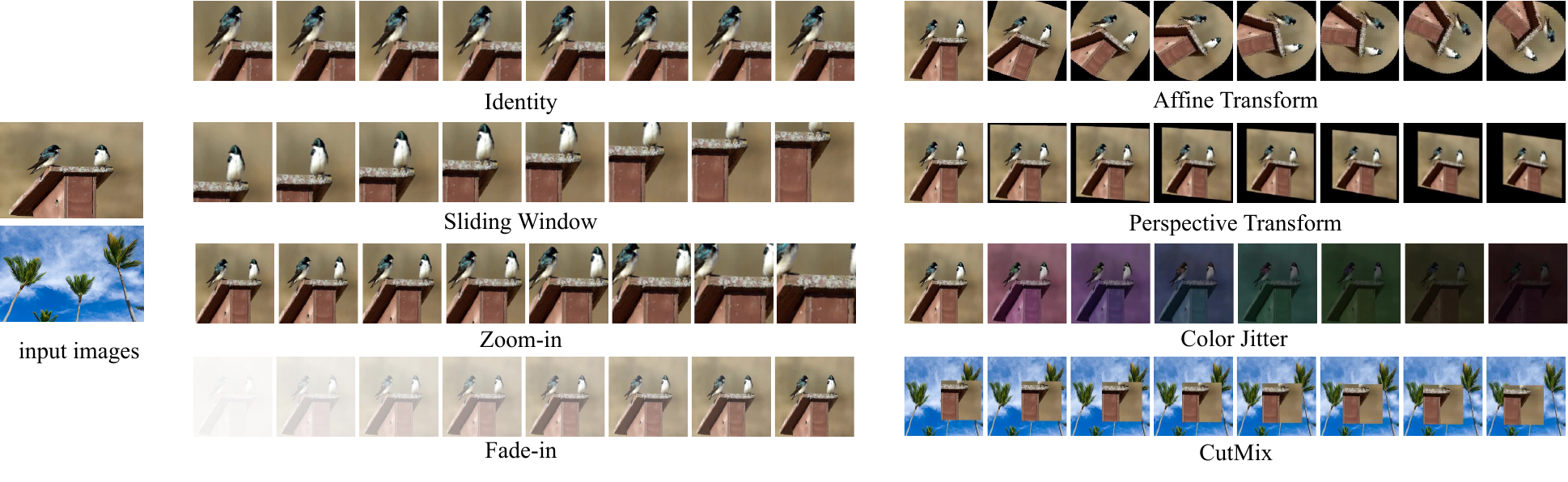}
    \vspace{-2mm}
    \caption{
    \textbf{Examples of pseudo-motion videos.}
    Images are sampled from PASS~\cite{asano2021pass}.
    For more examples of pseudo-motion videos,
    see the supplementary material.}
    \label{fig:example_pass}

    \vspace{-3mm}

\end{figure}






Figure~\ref{fig:example_pass} presents
examples of pseudo-motion videos generated by PMG.
Although the motions in these videos differ from real videos,
they exhibit a wide range of motion and appearance patterns.
Moreover, the clear correspondence of patches between frames
makes these pseudo-motion videos particularly well-suited for VideoMAE,
because it focuses on capturing low-level features
rather than high-level semantic features.
Notably, when pre-training VideoMAE using real videos,
we can use pseudo-motion videos generated from a frame within the videos
as a powerful form of data augmentation 
(we call this PMG Aug).
We demonstrate the effect of PMG Aug through experiments
(Refer to~\cref{sec:pmg_as_video_agumentation}).

\vspace{-3mm}
\subsection{Combination of Our Framework with Synthetic Images}
\label{sec:combination_with_synthetic_images}

Our framework enables the pre-training of video transformers using single images,
which are more accessible than real videos.
Additionally, our framework is applicable to synthetic images,
further reducing data collection costs and minimizing privacy
and other concerns associated with the use of real-world data.
We use the following synthetic image datasets for this purpose:
(i) FractalDB, generated based on fractal geometry~\cite{kataoka2020pre},
(ii) Visual Atom, created using sine waves~\cite{takashima2023visual},
(iii) Shaders1k, produced through OpenGL fragment shaders~\cite{baradad2022procedural}.
These datasets encompass a large volume and wide variety of images,
and have demonstrated to be as effective as real image datasets in the image recognition task.
By combining these synthetic images with our PMG module,
we can generate a wide variety of pseudo-motion videos,
enabling video transformers to learn effective spatio-temporal representations
as they would using real videos.

\vspace{-2mm}
\section{Experiments}
\label{sec:experimental_setup}

\noindent \textbf{Datasets.}
To evaluate the effectiveness of our framework,
we pre-train on pseudo-motion videos before fine-tuning and evaluating on various
action recognition datasets.
Following the SynAPT benchmark,
we use six action recognition datasets for fine-tuning and evaluation;
UCF101~\cite{soomro2012ucf101},
HMDB51~\cite{kuehne2011hmdb},
MiniSSV2~\cite{chen2021and} (a subset of Something-Something V2~\cite{goyal2017something}),
Diving48~\cite{li2018resound},
IkeaFA~\cite{toyer2017human},
and UAV-Human (UAV-H)~\cite{li2021uav}.
This benchmark is used to assess the transferability of our framework.
Additionally, we use Kinetics400 (K400)~\cite{kay2017kinetics}.
As an evaluation metric, we report the top-1 accuracy.

For pre-training, we adopt randomly sampled images from the following large-scale
image datasets:
ImageNet-1k (IN-1k)~\cite{deng2009imagenet},
PASS~\cite{asano2021pass},
FractalDB~\cite{kataoka2020pre},
Shaders1k~\cite{baradad2022procedural},
and Visual Atom~\cite{takashima2023visual}.
If the datasets have category annotations,
we sampled images so that the number of images of each category is the same.
Additionally, for comparison with ~\cite{huang2021self},
we randomly sample one frame of
a video and use it as an input image for generating pseudo-motion videos.

\noindent \textbf{Implementation Details.}
We conducted our experiments using 8 A100 GPUs.
Our training settings were mostly aligned with VideoMAE~\cite{tong2022videomae},
with a mask ratio of 0.75 and the number of epochs set to 2,000
unless otherwise noted
(See the supplementary material for details).
We used videos with 16 square frames (224 pixels in width).
For the model architecture,
we adopted a vanilla ViT~\cite{dosovitskiy2020image} as the backbone,
specifically the ViT-Base variant.

\vspace{-2mm}
\subsection{Ablation Studies}
\vspace{-1mm}
\label{sec:ablation}

\begin{table}[t]
\begin{minipage}[t!]{.45\textwidth}
\scriptsize
\caption{\footnotesize{\textbf{Comparison of different image augmentations.}}}
\label{tab:comparison_of_image_aug}

\vspace{-2mm}
\centering

\begin{tabular}{ccc}
\toprule[1.2pt]
Method                                                  & UCF101         & HMDB51 \\

\midrule[0.5pt]

Baseline (Identity)                                     & 72.7          & 35.6    \\
Sliding Window                                          & 75.1          & 40.5    \\
Zoom-in/out                                             & 81.2          & 44.5    \\
Fade-in/out                                             & 76.3          & 34.1    \\
Affine                                                  & 80.5          & 43.2    \\
Perspective                                              & \textbf{82.7} & \textbf{45.9}  \\
Color Jitter                                            & 76.2          & 38.7    \\
CutMix                                                  & 76.8          & 45.1    \\

\bottomrule[1.2pt]

\end{tabular}
\end{minipage}
\hspace{0.01\textwidth}
\begin{minipage}[t!]{.45\textwidth}
\scriptsize
\caption{\footnotesize{\textbf{Combination of image augmentations for PMG.}}}
\label{tab:comparison_of_image_aug_combination}
\vspace{-2mm}
\centering
\begin{tabular}{ccccc}
\toprule[1.2pt]

Zoom-in/out  & Affine     & Perspective  & CutMix      & HMDB51 \\

\midrule[0.5pt]

\checkmark   & \checkmark &              &             & \textbf{51.8}   \\
\checkmark   &            & \checkmark   &             & 45.2   \\
\checkmark   &            &              & \checkmark  & 41.4   \\
             & \checkmark & \checkmark   &             & 50.5   \\
             & \checkmark &              & \checkmark  & 47.1   \\
             &            & \checkmark   & \checkmark  & 44.2   \\
\checkmark   & \checkmark & \checkmark   &             & 49.0   \\
\checkmark   & \checkmark &              & \checkmark  & 49.4   \\
\checkmark   &            & \checkmark   & \checkmark  & 47.2   \\
             & \checkmark & \checkmark   & \checkmark  & 42.0   \\
\checkmark   & \checkmark & \checkmark   & \checkmark  & 47.9   \\

\bottomrule[1.2pt]

\end{tabular}

\end{minipage}

\vspace{-4mm}
\end{table}


\subsubsection{The effect of image augmentations.}
\label{sec:effect_of_image_augmentation}

First, we investigated the contribution of each image transformation
on VideoMAE pre-training.
We used HMDB51 and UCF101 for generating pseudo-motion videos for pre-training,
then used videos from the respective datasets to fine-tune the model.
\cref{tab:comparison_of_image_aug} reports the results when applying only a single variation
of image augmentation.
Videos generated by the Identity transformation serve as a baseline
because they do not contain any motion.
Compared to this baseline,
videos generated with Sliding Window, Zoom-in/out, Affine Transformation,
Perspective Transformation, and CutMix
improve the model's accuracy over the baseline.
Pseudo-motion videos generated with these transformations have corresponding patches between
frames, meaning that patches in one frame might slightly move but would still exist in the
subsequent frame.
Therefore, this supports our hypothesis that this characteristic aids the VideoMAE when learning
spatio-temporal features.

While videos generated with Fade-in/out and Color Jitter marginally improved performance
on UCF101, they did not do as well on HMDB51,
which is a motion-sensitive dataset~\cite{li2018resound}.
This suggests that videos made with these transformations are beneficial for capturing
spatial features but do not aid in capturing motion features.
Next, we experimentally determine the optimal set of image transformations $\phi$
from Sliding Window, Zoom-in/out, Affine Transformation,
Perspective Transformation, and CutMix.

\vspace{-3mm}
\subsubsection{The combination of image augmentations.}
\label{sec:combination_of_image_aug}

\cref{tab:comparison_of_image_aug_combination} compares the performance on HMDB51
when models are pre-trained with various combinations of image transformations.
It is observed that combining multiple image transformations improves the model's performance.
This indicates that the model can effectively learn as long as there is sufficient diversity,
even if the motion patterns in pseudo-motion videos differ from those in real videos.
However, combining more image transformations did not necessarily yield better results.
In particular, in most cases where we applied CutMix, the accuracy decreased.
We hypothesize that this is due to the non-continuous nature of CutMix videos.
From this point on, we will use Zoom-in/out and Affine Transformation
as the set of image transformations $\phi$.
Further discussion on the failure cases of pre-training with these pseudo-motion videos
is provided in the supplementary material.

\vspace{-3mm}
\subsubsection{The efficacy of video-level augmentations.}

To further enhance the diversity of videos,
we applied video-level augmentation to the generated pseudo-motion videos.
We examined two methods: Mixup~\cite{zhang2017mixup} and VideoMix~\cite{yun2020videomix}.
~\cref{tab:video_level_aug} demonstrates that video-level augmentation, especially Mixup,
significantly contributes to performance improvement.
This is because both video augmentations diversify pseudo-motion videos,
resulting in better performance.
Pre-training with VideoMix results in lower accuracy compared to Mixup
because the videos generated by VideoMix have non-continuous regions
like CutMix, as discussed in~\cref{sec:combination_of_image_aug},
From here, we will utilize Mixup in our experiments.

\begin{table}[t]
\begin{minipage}{.49\textwidth}
\centering
\scriptsize
\caption{
\textbf{Effects of video-level augmentation.}
}
\label{tab:video_level_aug}
\vspace{-2mm}
\begin{tabular}{cccc}
\toprule[1.2pt]

\multicolumn{2}{c}{Video Augmentaiton}         & \multicolumn{2}{c}{Dataset} \\
Mixup                     & VideoMix            & HMDB51     & UCF101 \\
\midrule[0.5pt]

                          &                     & 51.8       & 83.8   \\
\checkmark                &                     & \textbf{55.9} & \textbf{87.3} \\
                          & \checkmark          & 53.0       & 85.2   \\

\bottomrule[1.2pt]

\end{tabular}

\end{minipage}
\hspace{0.01\textwidth}
\begin{minipage}{.49\textwidth}
\scriptsize
\centering
\caption{\textbf{Transferability from other video datasets.}}
\label{tab:transferability}
\vspace{-2mm}
\begin{tabular}{ccc}
\toprule[1.2pt]

Pre-training & Fine-tuning & Top1               \\
\midrule[0.5pt]
HMDB51      & \multirow{2}{*}{HMDB51} & 55.9  \\
UCF101      &                         & 56.7  \\
\midrule[0.5pt]
UCF101      & \multirow{2}{*}{UCF101} & 87.3   \\
HMDB51      &                         & 85.5  \\
\bottomrule[1.2pt]

\end{tabular}
\end{minipage}

\vspace{-3mm}
\end{table}


\vspace{-2mm}
\subsection{Transferability of Our Framework}
\vspace{-1mm}

\subsubsection{Transferability from other video datasets.}
\label{sec:transferability}
To verify the transferability of our framework,
we conducted experiments by pre-training models with pseudo-motion videos generated from 
frames in HMDB51 and then fine-tuning on
UCF101 (hereafter, we refer to this as HMDB51 $\rightarrow$ UCF101),
and then vice versa (UCF101 $\rightarrow$ HMDB51).
\cref{tab:transferability} shows the results.
Comparing the accuracy when pre-training on different datasets, the difference
is marginal.
This suggests that our framework learns robust features that are not
domain-specific.
Furthermore, this appeals that our framework can effectively pre-train
models even when using image datasets instead, such as ImageNet and PASS.

\vspace{-2mm}
\subsubsection{Transferability from real image datasets.}
\label{sec:experiment_with_image_datasets.}
\vspace{-1mm}

In our previous experiments,
we used samples with similar visual cues between pre-training and fine-tuning,
namely the semantic information including objects and people.
To further assess the transferability of our framework,
we conducted pre-training on the ImageNet-1k and PASS,
which are in different domains compared to the fine-tuning datasets (UCF101 and HMDB51).
As detailed in~\cref{tab:imagenet_pass_results},
pre-training using ImageNet and PASS 
achieved comparable performance to when pre-training with the same
datasets that are used when fine-tuning.
Note that PASS does not include any human images.
Therefore, the semantic information within pre-training datasets are not
a must for effective pre-training of VideoMAE.
Moreover, increasing the number of images scaled the performance.
These experimental results suggest that for VideoMAE,
the diversity of the data is more crucial
than domain-specific information like human motion or visual cues.

\vspace{-3mm}
\subsubsection{Transferability from synthetic images.}
\label{sec:experiments_with_synthetic_data}

\begin{table}[t]
\begin{minipage}{.48\textwidth}
\centering
\scriptsize
\caption{\textbf{Pre-training with ImageNet and PASS.}
The term ’FT data’ indicates that the
datasets used for pre-training are identical to those used in fine-tuning.
}
\label{tab:imagenet_pass_results}

\vspace{-2mm}
\begin{tabular}{cccc}
\toprule[1.2pt]
\multicolumn{2}{c}{Pre-training} & \multicolumn{2}{c}{Downstream task} \\
Dataset       & \#data          & UCF101 & HMDB51 \\
\midrule[0.5pt]
FT data       & -               & 87.3   & 55.9   \\
\midrule[0.5pt]
ImageNet      & 10,000          & 87.4   & 58.0   \\
ImageNet      & 100,000         & 89.2   & 59.2       \\
\midrule[0.5pt]
PASS          & 10,000          & 87.6   & 58.3   \\
PASS          & 100,000         & \textbf{89.3}   & \textbf{60.0}   \\
\bottomrule[1.2pt]
\end{tabular}
\end{minipage}
\hspace{0.01\textwidth}
\begin{minipage}{.49\textwidth}
\scriptsize
\centering
\caption{\textbf{Pre-training on synthetic image datasets.}}
\label{tab:synthetic_datasets_results}
\vspace{-2mm}
\begin{tabular}{cccc}
\toprule[1.2pt]
\multicolumn{2}{c}{Pre-training Setting} & \multicolumn{2}{c}{Downstream task}      \\
Dataset                      & \#Data     & UCF101       & HMDB51 \\
\midrule[0.5pt]
FT data & -               & 87.3   & 55.9   \\
\midrule[0.5pt]
\multirow{2}{*}{FractalDB}   & 10,000     & 77.6         & 42.8    \\
                             & 100,000    & 78.1         & 41.1    \\
\midrule[0.5pt]
\multirow{2}{*}{Shaders1k}   & 10,000     & 88.4         & 57.6    \\
                             & 100,000    & \textbf{89.6}& \textbf{59.7} \\
\midrule[0.5pt]
\multirow{2}{*}{Visual Atom} & 10,000     & 83.5        & 48.9    \\
                             & 100,000    & 82.6        & 48.2     \\
\bottomrule[1.2pt]
\end{tabular}
\end{minipage}

\vspace{-3mm}
\end{table}


We then pre-trained on synthetic image datasets using our framework
to verify that spatio-temporal features can be effectively learnt
from synthetic images, which present completely different visual cues
compared to our target action recognition datasets.
For synthetic image datasets, we used FractalDB, Shaders1k, and Visual Atom.
Herein, we used 10k/100k images sampled from each dataset.
\cref{tab:synthetic_datasets_results} shows
the performance on UCF101 and HMDB51
when pre-training on diverse synthetic datasets,
including FractalDB, Shaders1k, and VisualAtom.
Note that pre-training with Shaders1k achieved
comparable results to pre-training with real images, where
pre-training with FractalDB and Visual Atom lead to subpar performance.
This denotes that the model struggles to correlate patches between
frames of pseudo-motion videos generated from FractalDB and Visual Atom, thus
failing to capture robust low-level features.
On the other hand, images in Shaders1k have distinctive patches that can be
correlated before and after transformations, which supports the model when
capturing low-level features.
This indicates that our framework can successfully replace the need for real
data when pre-training the model, as long as synthetic videos have patches that
can be tracked between frames.
Thus, when using our framework, challenges related to real datasets such as privacy
and license infringement are nonexistent.

\vspace{-3mm}
\subsection{Effect of the Number of Epochs, Data, and Categories in Image Datasets for Pre-training}

\cref{fig:epoch_vs_acc} shows the relationship between the number of epochs
of pre-training and accuracy on HMDB51.
For generating videos for pre-training, we used 10k images from Shaders1k and a frame from
each of the 3k videos in HMDB51.
In both datasets, the model performance improved over epochs and
the difference of accuracy gradually decreased.
Because our PMG allows for the generation of diverse videos,
even if we have a small amount of data for pre-training,
it is possible to improve performance by increasing the number of iterations.

\begin{figure}[t]
\centering
\footnotesize

\begin{subfigure}[b]{0.32\textwidth}
\centering
\includegraphics[width=\textwidth]{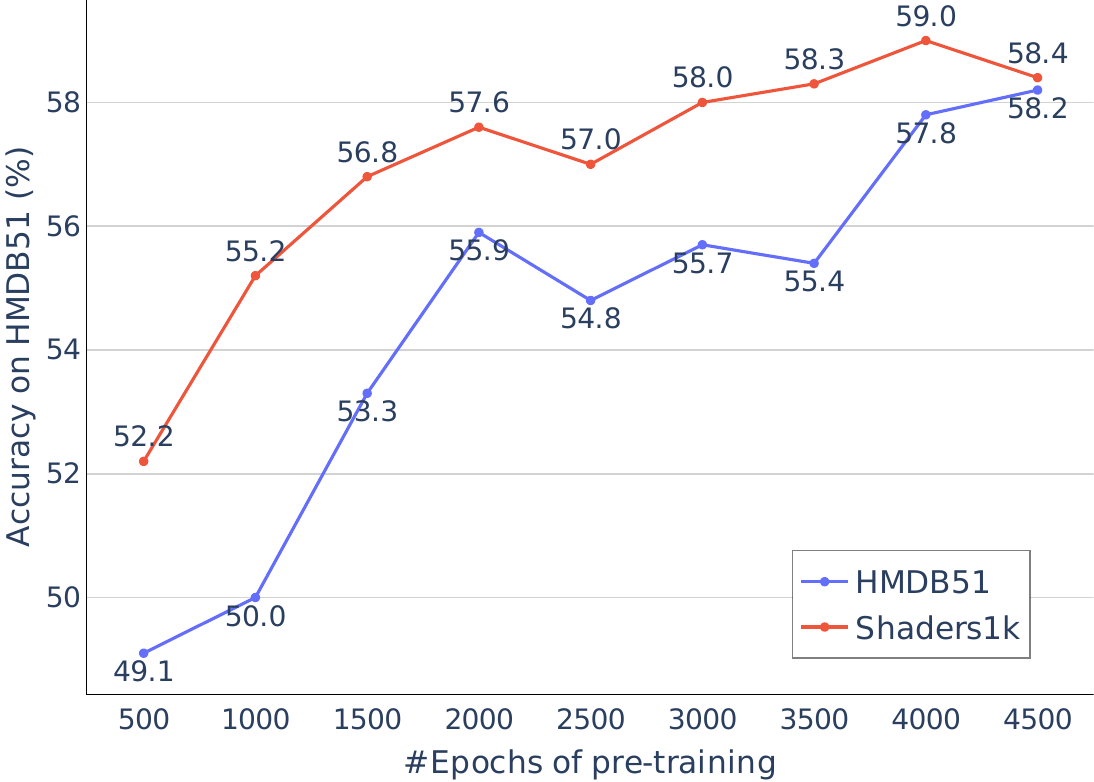}
\caption{Pre-training epochs vs accuracy on HMDB51}
\label{fig:epoch_vs_acc}
\end{subfigure}
\hfill
\begin{subfigure}[b]{0.32\textwidth}
\centering
\includegraphics[width=\textwidth]{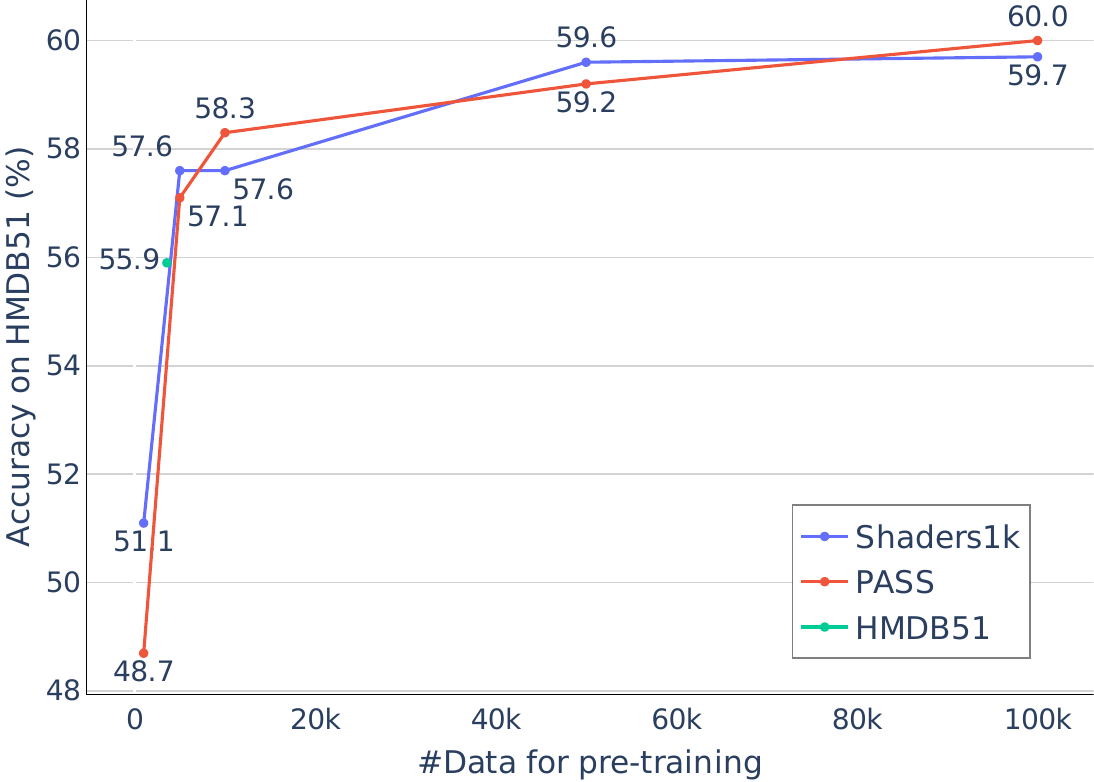}
\caption{The number of data vs accuracy on HMDB51}
\label{fig:num_data_vs_acc}
\end{subfigure}
\hfill
\begin{subfigure}[b]{0.32\textwidth}
\centering
\includegraphics[width=\textwidth]{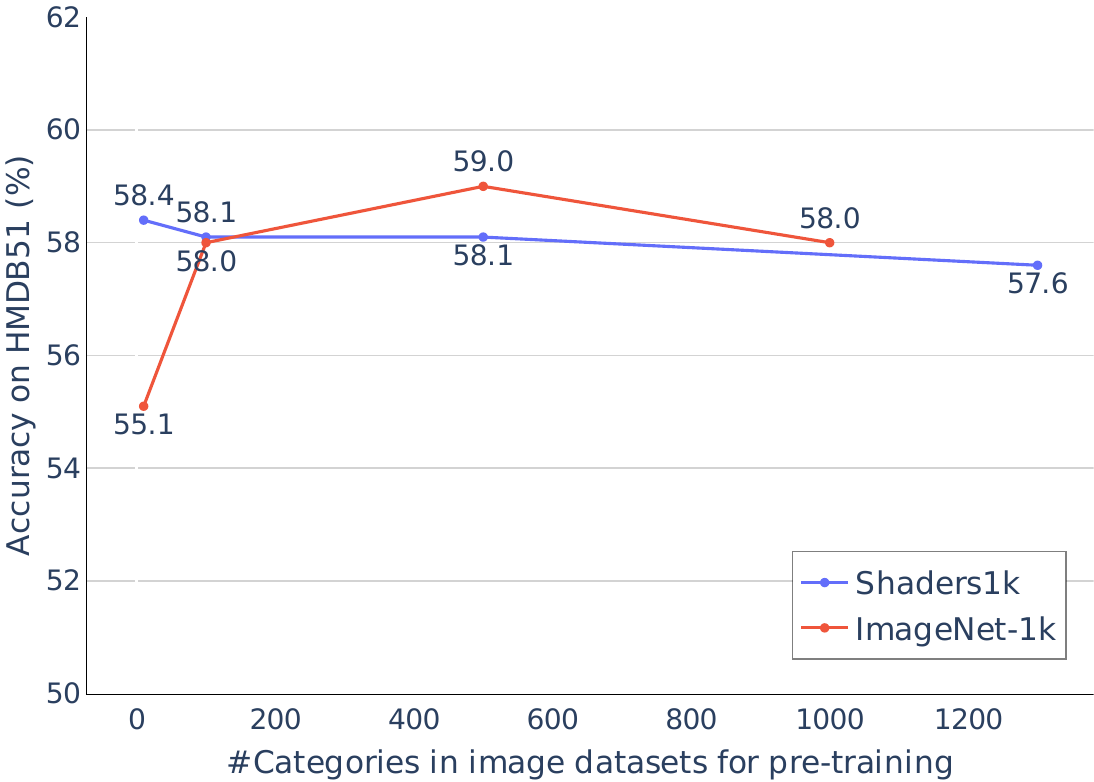}
\caption{The number of categories vs accuracy on HMDB51}
\label{fig:class_vs_acc}
\end{subfigure}
\vspace{-1mm}
\caption{\textbf{Effect of the number of epochs, data, categories.}}
\vspace{-2mm}
\label{fig:ablation_graph}
\end{figure}

\cref{fig:num_data_vs_acc} presents the accuracy transition
when the number of pre-training samples is varied among $\{1k, 5k, 10k, 50k, 100k\}$.
Our framework shows improvement as the number of data increased.
Because we use only a small subset from PASS and Shaders1k,
there is potential for more substantial performance improvement
when generating from all images.

Based on the results of the previous experiment,
we hypothesized that
performance can be further enhanced
by increasing the diversity of samples in pre-training image datasets.
\cref{fig:class_vs_acc} shows the relationship between
the number of categories in the pre-training datasets we use
and the classification performance on HMDB51.
We set the number of training samples to 10k images, using the IN-1k and Shaders1k datasets.
For IN-1k, the accuracy seems to saturate after raising the diversity to more than 50 classes.
For Shaders1k, the accuracy was almost the same
even when the number of categories increased.
This suggests our framework scales with having more data samples,
but does not require semantic diversity within the samples.
These results support the fact that VideoMAE learns low-level features like the
correspondence of patches between frames, rather than semantic information like categories of objects
displayed.

\begin{table}[t]
\begin{minipage}{.47\textwidth}
\centering
\scriptsize
\caption{
    \textbf{Effectiveness of our PMG as a video augmentation method on HMDB51 and UCF101.}
    $^\ast$ results from~\cite{tong2022videomae}.
}
\label{tab:pvg_as_aug}
\vspace{-2mm}
\begin{tabular}{cc|cc}
\toprule[1.2pt]
\multicolumn{2}{c|}{Pre-training data}& \multicolumn{2}{c}{Downstream tasks}     \\
~real video~         & ~pseudo motion & HMDB51~                  & UCF101         \\
\midrule[0.5pt]
\checkmark           &                & 62.6$^\ast$              &  91.3$^\ast$   \\
                     & \checkmark     & 55.9                     &  87.3          \\
\checkmark           & \checkmark     & \textbf{64.6}            &  \textbf{92.2} \\
\bottomrule[1.2pt]

\end{tabular}
\end{minipage}
\hspace{0.01\textwidth}
\begin{minipage}{.52\textwidth}
\scriptsize
\centering
\caption{
    \textbf{Comparison of each combination with real videos and pseudo-motion videos.}
    $^\ast$ Sources of PMG Aug.
}
\label{tab:comparison_of_dataset_combination}
\vspace{-2mm}

\begin{tabular}{cccc|c}
\toprule[1.2pt]
\multicolumn{4}{c|}{Pre-training data}                           & \multirow{2}{*}{~HMDB51} \\
Videos            &~frames$^\ast$~& ~PASS$^\ast$ &~Shaders1k$^\ast$      &                 \\
\midrule[0.5pt]
\checkmark        &               &              &                & 62.6            \\ 
\checkmark        & \checkmark    &              &                & 64.6            \\ 
\checkmark        & \checkmark    & \checkmark   &                & \textbf{68.0}   \\ 
\checkmark        & \checkmark    &              & \checkmark     & 67.0            \\
\checkmark        & \checkmark    & \checkmark   & \checkmark     & 67.9             \\
\bottomrule[1.2pt]
\end{tabular}

\end{minipage}
\vspace{-4mm}
\end{table}


\vspace{-2mm}
\subsection{PMG as Video Augmentation on Pre-training}
\vspace{-1mm}
\label{sec:pmg_as_video_agumentation}

\subsubsection{Pre-training with both real videos and pseudo-motion videos.}

As we have verified so far,
our framework enables efficient pre-training with static images.
This suggests that our proposed PMG can be also used
as a data augmentation method during pre-training.
\cref{tab:pvg_as_aug} compares the performance
when pre-training solely with real videos
and when pre-training with both real and pseudo-motion videos.
Notably, PMG Aug boosted model accuracy by up to 2\%.
This suggests that synthetic motion, despite its differences to real video motion,
unintuitively contributes to the model's performance by increasing diversity.

\begin{table*}[t]
\centering
\scriptsize
\caption{
    \textbf{Comparison with existing methods on HMDB51, UCF101, and Diving48.}
    RV = Real Videos, SV = Synthetic Videos, RI = Real Images, SI = Synthetic Images,
    SP = Supervised Pre-training,
    FT data = Fine-tuning data.
    $^\dag$ Results in our replicated experiments.
    $^\ddag$ Reported in~\cite{kim2022transferable}.
    $^\ast$ Herein, we refer to a combination of
    ElderSim~\cite{hwang2021eldersim},
    SURREACT~\cite{varol2021synthetic},
    and PHAV~\cite{roberto2017procedural}
    as SynAPT, as proposed in ~\cite{kim2022transferable}.
    $^\S$ we report only the number of videos in SynAPT.
}
\label{tab:comparison_with_sota}

\vspace{-2mm}
\begin{tabular}{cccc|ccc}
\toprule[1.2pt]
\multirow{2}{*}{Method}                         & \multicolumn{3}{c|}{Pre-training Setting}                                            & \multicolumn{3}{c}{Downstream Tasks}      \\
                                                & Dataset                                                  & Data Source & \#Data & UCF101         & HMDB51       & Diving48 \\
\midrule[0.5pt]
from scratch (ViT-B)                            & -                                                        & -           & -      & 51.4           & 18.0         & 17.9$^\dag$ \\
MoCo v3 (ViT-B)~\cite{chen2021and}              & FT data                                                  & RV          & -      & 81.7           & 39.2         & -        \\
VideoMAE (ViT-B)~\cite{tong2022videomae}        & FT data                                                  & RV          & -      & 91.3           & 62.6         & 79.3$^\dag$ \\
VideoMAE (ViT-B)~\cite{tong2022videomae}        & Kinetics400                                              & RV          & 260k   & 96.1           & 73.3         & -        \\
VideoMAE (ViT-B)$^\dag$                         & VPN~\cite{kataoka2022spatiotemporal}                     & RV          & 10k    & 64.9           & 30.3         & 17.5    \\
3D-ResNet50~\cite{hara2018can}                  & VPN~\cite{kataoka2022spatiotemporal}                     & SV          & 28k    & 49.9           & 23.0         & -        \\
3D-ResNet50~\cite{hara2018can}                  & VPN$\rightarrow$Kinetics400                              & RV + SV     & 280k   & 89.9           & 61.8         & -        \\
TimeSformer~\cite{bertasius2021space}$^\ddag$    & IN-21k$\rightarrow$SynAPT$^\ast$                        & RI + SV     & 150k$^\S$ & 89.0           & 54.4         & 44.9     \\
PPMA~\cite{zhong2023learning}               & NH-Kinetics+SynAPT$^\ast$                                & RV + SV     & 300k   & 92.5           & 71.2         & 64.0     \\
\midrule[0.5pt]
\midrule[0.5pt]
MoSI (R-2D3D)~\cite{huang2021self}              & FT data                                                  & RI          & -      & 71.8           & 47.0         & -        \\
MoSI (R(2+1)D)~\cite{huang2021self}             & FT data                                                  & RI          & -      & 82.8           & 51.8         & -        \\
MoSI (ViT-B)$^\dag$                             & FT data                                                  & RI          & -      & 48.0           & 27.3         & 14.2 \\
SP (ViT-B)$^\dag$                               & IN-21k                                                   & RI          & 14M    & 71.9           & 34.0         & 34.2         \\
SP (ViT-B)$^\dag$                               & ExFractalDB-21k~\cite{kataoka2022replacing}              & SI          & 21M    & 61.5           & 20.8         & 28.0         \\
SP (ViT-B)$^\dag$                               & VisualAtom-21k                                           & SI          & 21M    & 58.9           & 20.3         & 21.4         \\
\midrule[0.5pt]
\multirow{3}{*}{Ours (ViT-B)}            & frames from FT data                                      & RI          & -      & 87.3           & 55.9         & 68.3      \\
                                         & PASS                                                     & RI          & 100k   & 89.3           & \textbf{60.0}& 69.2    \\
                                         & Shaders1k                                                & SI          & 100k   & \textbf{89.4}  & 59.7         & \textbf{72.3} \\
\bottomrule[1.2pt]
\end{tabular}
\vspace{-4mm}
\end{table*}


\vspace{-3mm}
\subsubsection{Can we combine image datasets with video datasets to train our framework?}
Next, we use image datasets as well as sources of PMG Aug during pre-training.
For the image datasets (PASS and Shaders1k), we randomly sampled 10k images as input.
\cref{tab:comparison_of_dataset_combination} compares
the performance of the models pre-trained on HMDB51,
PASS, and Shaders1k.
The results show that using both image and video datasets
improved the model's performance.
Particularly, the combination of HMDB51 and PASS
enhanced the accuracy on HMDB51 by 5.4\%
compared to pre-training with only real videos.
This indicates that using PMG Aug resolves the problem of insufficient data quantity
during VideoMAE pre-training.

\vspace{-2mm}
\subsection{Comparison to Existing Methods}
\label{sec:compariton_with_sota}
\vspace{-2mm}

\subsubsection{Comparison to methods using HMDB51, UCF101 and Diving48.}

The upper part of~\cref{tab:comparison_with_sota} presents 
the performance of existing works which pre-train using the
HMDB51, UCF101, and Diving48 datasets.
Existing methods like 3D-ResNet with VPN~\cite{kataoka2022spatiotemporal},
TimeSformer with SynAPT, and PPMA
have improved model performance compared to training the model from scratch.
However, they still require real data, causing issues as mentioned.
In contrast, our framework,
despite using fewer samples which are also synthetic,
achieves comparable performance on UCF101
and better performance on Diving48.

We also compared with pre-training methods which only use static images
(the lower part of~\cref{tab:comparison_with_sota}).
MoSI works on CNN-based architectures,
but it fails to pre-train a ViT model because of the lack of diversity in generated
videos.
Supervised pre-training (SP) on IN-21k, ExFractalDB-21k~\cite{kataoka2022replacing}
and VisualAtom-21k
slightly improves the performance in comparison with 'from scratch'.
However, our framework significantly surpasses that performance in both settings,
when using real images and when using synthetic images.

Note that VideoMAE pre-trained with VPN has low accuracy on downstream classification tasks,
which suggests that VPN does not work well with VideoMAE when learning spatio-temporal features.
We consider this is because VPN videos have temporal continuity,
but do not possess clear correspondence of patches between frames
(e.g. edges are ambiguous, and regions suddenly disappear or appear).
We believe this characteristic is key for effective VideoMAE pre-training.
In~\cref{sec:what_videomae_learns},
we further experiment to support this hypothesis.

\begin{table}[t]

\centering
\scriptsize

\caption{
    \textbf{Results on SynAPT benchmark.}
    $^\dag$ Results reported in~\cite{kim2022transferable}.
}
\label{tab:synapt}
\vspace{-1mm}
\begin{tabular}{cccc|cccccc}
\toprule[1.2pt]
\multirow{2}{*}{Method}       & \multicolumn{3}{c|}{Pre-training}                                         & \multicolumn{6}{c}{Downstream Tasks}                   \\
                              & Dataset                                                 & \#data    & ~labels & UCF101 & HMDB51 & MiniSSV2 & Diving48 & IkeaFA & UAV-H \\
\midrule[0.5pt]
TimeSformer$^\dag$            & \begin{tabular}{c}IN-21k\\+Synthetic\end{tabular}       & 150k       & \checkmark & 89.0   & 54.4 & 51.1          & 44.9     & 63.6     & 25.9      \\
\midrule[0.5pt]
PPMA~\cite{zhong2023learning} & \begin{tabular}{c}NH-Kinetics\\+Synthetic\end{tabular}  & 300k       & \checkmark & \textbf{92.5} & \textbf{71.2} & 67.8      & 64.0     & \textbf{67.9}   & 38.5      \\
\midrule[0.5pt]
Ours                 & no extra data                                             & -                 &   & 87.3   & 55.9   & \textbf{69.0}        & 68.3     & 61.4   & 36.8      \\
\midrule[0.5pt]
Ours                 & Shaders1k                                                 & 100k              &   & 89.4   & 59.7   & 68.3 & \textbf{72.3}  & 60.7   & \textbf{40.0}        \\
\bottomrule[1.2pt]

\vspace{-4mm}
\end{tabular}
\end{table}


\begin{table}[t]
\begin{minipage}{.52\textwidth}
\centering
\scriptsize
\caption{
\textbf{Results on K400.}
$^\dag$ Results from ~\cite{tong2022videomae}.
We use 100k images from Shaders1k.}
\vspace{-2mm}
\label{tab:k400_ssv2}
\begin{tabular}{cccc}
\toprule[1.2pt]
\multirow{2}{*}{Method} & Pre-training data & \multicolumn{2}{c}{Kinetics400}  \\
                        & Data              & Acc@1   & Acc@5 \\
\midrule[0.5pt]
from scratch$^\dag$            & -                  & 68.8     & -    \\
VideoMAE$^\dag$                & K400              & \textbf{81.5} & \textbf{95.1}  \\
\midrule[0.5pt]
\multirow{2}{*}{Ours}    & frames from K400  & 74.8    & 92.0  \\
                         & Shaders1k         & 74.7    & 91.9   \\
                         
\bottomrule[1.2pt]
\end{tabular}
\end{minipage}
\hfill
\begin{minipage}{.47\textwidth}
\scriptsize
\centering
\caption{\textbf{Comparison of accuracy on HMDB51 and UCF101 when using subsets grouped by frame difference.}}
\vspace{-1mm}
\begin{tabular}{ccc}
\toprule[1.2pt]
Frame difference & HMDB51        & UCF101 \\
\midrule[0.5pt]
(i) Large           & 32.9          & 68.7   \\
(ii) Medium         & \textbf{33.5} & \textbf{71.0}   \\
(iii) Small         & 32.2          & 69.3   \\
\bottomrule[1.2pt]
\end{tabular}
\label{tab:what_video_learns1}
\end{minipage}

\vspace{-3mm}
\end{table}


\vspace{-3mm}
\subsubsection{Comparison on SynAPT benchmark.}
Following the SynAPT benchmark~\cite{kim2022transferable},
we evaluate using the following six datasets:
UCF101, HMDB51, MiniSSV2, Diving48, IkeaFA, and UAV-H.
\cref{tab:synapt} presents the results.
Using only synthetic images,
our proposed framework partially surpasses some of the results of existing works utilizing
real videos and action labels.
Our framework is inferior to PPMA on UCF101, HMDB51, and IkeaFA.
This is because these datasets have less data than others.
PPMA leverages the 150 action labels in the video datasets for pre-training,
therefore having the advantage of learning action features from a small number of videos
during fine-tuning.
On the other hand, our framework, not having these labels beforehand, struggled to learn
meaningful features with fewer labeled data.
However, our framework shows better performance
on less biased datasets like MiniSSV2, Diving48, and UAV-H.
This suggests that scene and object biases are mitigated when using our generated synthetic
videos.

\vspace{-3mm}
\subsubsection{Results on K400.}
\label{sec:k400}

~\cref{tab:k400_ssv2} shows the comparison of our framework with VideoMAE on K400.
Although our framework outperforms the model 'from scratch',
it falls short of the performance of VideoMAE with real videos.
This shortfall is attributed to the limited diversity of pseudo-motion videos
generated by PMG,
especially when compared to the vast variety found in large-scale datasets.
We understand our shortcoming here, but increasing the diversity of generated videos
may close this gap.

\begin{figure}[t]
\centering
\footnotesize
\scriptsize

\begin{subfigure}[b]{0.48\textwidth}
\centering
\includegraphics[width=\textwidth]{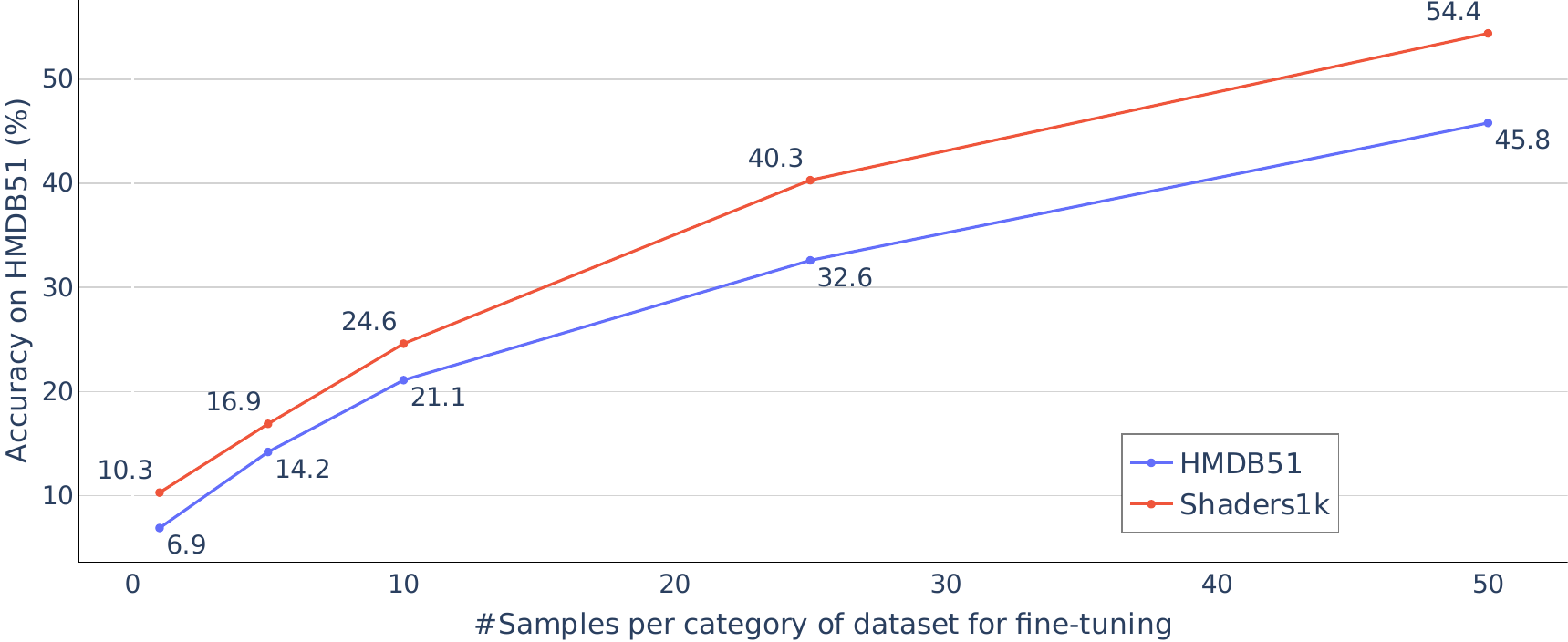}
\caption{Accuracy transition on HMDB51}
\label{fig:epoch_vs_acc}
\end{subfigure}
\hfill
\begin{subfigure}[b]{0.48\textwidth}
\centering
\includegraphics[width=\textwidth]{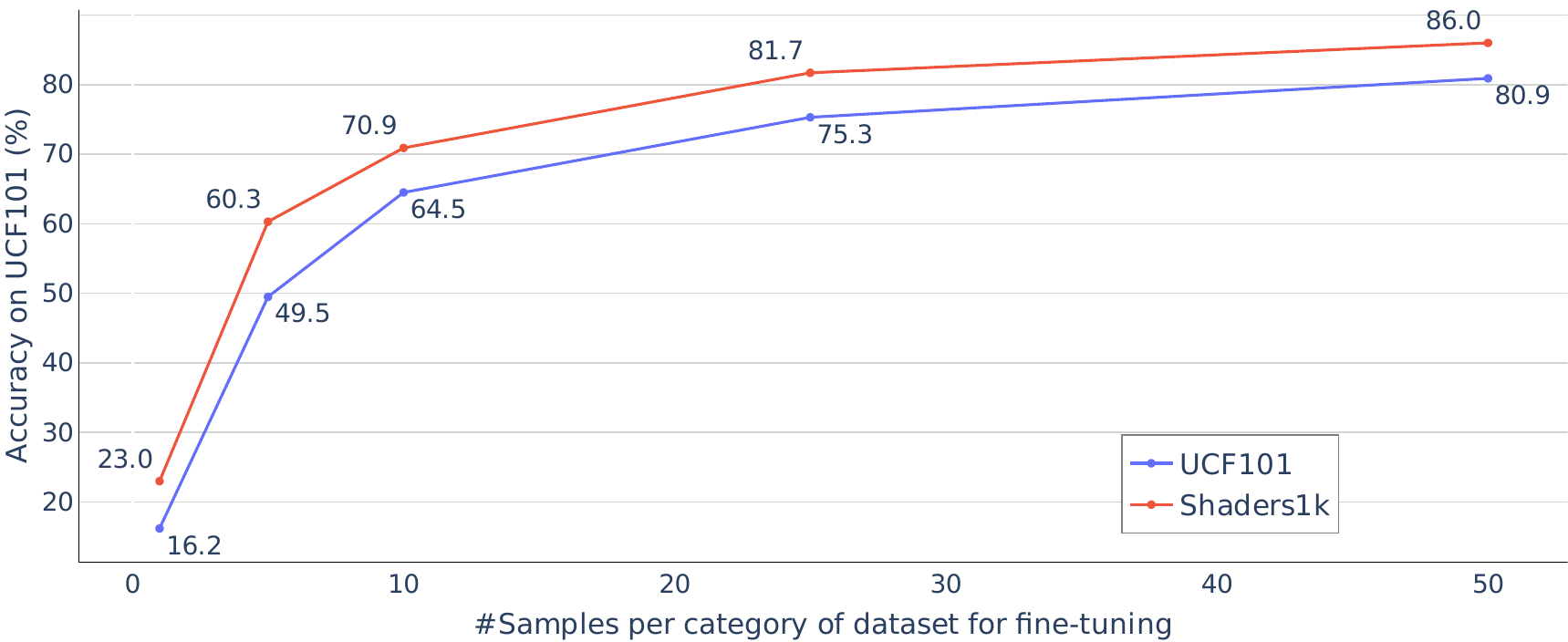}
\caption{Accuracy transition on UCF101}
\label{fig:class_vs_acc}
\end{subfigure}
\vspace{-1mm}

\caption{\textbf{Performance when the number of video data for finetuning is limited.}}
\vspace{-3mm}
\label{fig:limited_data}
\end{figure}

\vspace{-3mm}
\subsection{Performance When the Number Data for Fine-tuning is Limited}
\vspace{-1mm}

In previous experiments,
the full set of video datasets for fine-tuning was available.
Under these conditions, pre-training with all the videos for fine-tuning
yielded better performance than our framework.
However, for video datasets, there is often a limited amount of training samples to fine-tune with.
To assess the effectiveness of our framework in such cases,
we sampled $\{1, 5, 10, 25, 50\}$ videos per category from HMDB51 and UCF101, respectively,
and compared the performance of our framework with VideoMAE using real videos.
\cref{fig:limited_data} presents the results.
The model pre-trained by our framework
shows higher performance compared to the model pre-trained by VideoMAE using real data.
This underscores the efficacy of our framework where the available data is limited.

\vspace{-3mm}
\subsection{What Does VideoMAE Learn from Pre-training with Videos?}
\vspace{-1mm}
\label{sec:what_videomae_learns}

Finally, to support our hypothesis that
VideoMAE learns the correspondence of patches between frames,
we conducted a simple experiment.
Here, we assume that a larger frame difference in a video makes
it difficult to capture this correspondence,
for instance, due to extreme camera motion.
Based on this, we made three subsets from HMDB51 and UCF101 depending on the frame difference;
(i) videos having the top 50\% average frame difference 
(ii) videos ranging from the 25th to the 75th percentile in
average frame difference,
(iii) videos having the bottom 50\% average frame difference.
We then use each of these subsets for pre-training, then fine-tune on the full set.
The results are shown in~\cref{tab:what_video_learns1}.
Models that are pre-trained on (i) and (iii)
performed worse than those pre-trained on (ii).
This lends support to our hypothesis regarding what VideoMAE learns.

\vspace{-2mm}
\section{Conclusion}
\vspace{-1mm}

In this paper, we introduced a self-supervised framework
for pre-training video transformers solely with synthetic images.
Our framework eliminates the costs associated with collecting video data
and addresses concerns related to privacy, licensing, and biases inherent in real data.
Our experiments have demonstrated that our framework 
not only outperforms existing pre-training methods with static images
but also partially outperforms existing works with synthetic videos.
Further analysis unveiled segments of what masked autoencoders learn from videos.

\vspace{2mm}
\noindent\textbf{Limitations}
Our framework is inferior to pre-training with large-scale datasets like K400.
We consider this to be due to a lack of fine-grained motion patterns
compared to real videos.
Our framework depends on hand-crafted image transformations and
applies them to images globally,
pseudo-motion videos do not have flexible motion patterns.
Additionally, our framework does not learn high-level semantic features,
because we utilized VideoMAE's focus on capturing low-level features.
Therefore, it is challenging to extend our framework to
other tasks like video-text retrieval,
without additional training or extra labeled data.


%
%
\bibliographystyle{splncs04}
\bibliography{egbib}

\newpage
\appendix

\setcounter{figure}{\value{figure}}
\setcounter{table}{\value{table}}

\section*{Overview of Supplementary Material}

In this supplementary material,
we provide more details on our framework
and analyses of our experiments
with respect to the following points:

\begin{itemize}
    \item Details on video datasets (\cref{sec:dataset})
    \item Implementation details (\cref{sec:implementation})
    \item Pseudo-code of Pseudo Motion Generator (PMG) (\cref{sec:pmg})
    \item Parameters of image augmentations in PMG (\cref{sec:parameter})
    \item Examples of pseudo-motion videos generated by PMG (\cref{sec:example})
    \item Quantitative results of our framework (\cref{sec:quantitative})
    \item Failure cases (\cref{sec:failure_cases})
    \item Linear probing (\cref{sec:linear_probing})
\end{itemize}

\section{Details on Video Datasets}
\label{sec:dataset}

In our experiments,
we use seven datasets to evaluate the effectiveness of our framework;
UCF101~\cite{soomro2012ucf101},
HMDB51~\cite{kuehne2011hmdb},
MiniSSV2~\cite{chen2021and},
Diving48~\cite{li2018resound},
IkeaFA~\cite{toyer2017human},
UAV-Human (UAV-H)~\cite{li2021uav},
and Kinetics400 (K400)~\cite{kay2017kinetics}.
The first six datasets are included in the SynAPT benchmark~\cite{kim2022transferable}
We conducted our experiments following its setup.
Herein, we provide an overview of the datasets used in this study.

\noindent \textbf{UCF101~\cite{soomro2012ucf101}:}
This dataset features approximately 13,000 videos classified into 101 categories of actions.
These categories are segmented into five groups:
(i) Human-object Interaction (e.g., Juggling Balls),
(ii) Body-Motion Only (e.g., Push Ups),
(iii) Human-Human Interaction (e.g., Head Massage),
(iv) Playing Musical Instruments (e.g., Drumming),
and (v) Sports (e.g., Archery).

\noindent\textbf{HMDB51~\cite{kuehne2011hmdb}:}
Comprising roughly 6,000 video clips sourced from both movies and YouTube,
this dataset is annotated across 51 action categories.
These categories encompass five action types:
(i) general facial actions (e.g. smile),
(ii) facial actions with object manipulation (e.g. eat),
(iii) general body movements (e.g. jump),
(iv) body movements with object interaction (e.g. kick ball),
(v) body movements for human interaction (e.g. punch).

\noindent \textbf{MiniSSV2~\cite{chen2021and}:}
MiniSSV2~\cite{kim2022transferable} is a subset of Something-Something V2 (SSV2)~\cite{goyal2017something},
which encompasses over 220,000 video clips with 174 action classes.
MiniSSV2 contains just half of the original action categories,
with 87 randomly selected labels.
The total number of videos is approximately 93,000 videos.
Actions in this dataset are basic interactions with everyday objects,
defined via caption templates like "Moving something up" or "Covering something with something".

\noindent \textbf{Diving48~\cite{li2018resound}:}
Dedicated to competitive diving,
this dataset consists of about 18,000 videos categorized into 48 distinct types of diving actions.
All videos in Diving48 exhibit consistent background and object characteristics.
Therefore, this dataset is often used to evaluate how the models capture motion information.

\noindent \textbf{IkeaFA~\cite{toyer2017human}:}
Ikea Furniture Assembly (IkeaFA) offers 111 video clips,
each lasting between 2 to 4 minutes, accumulating roughly 480,000 frames.
This dataset consists of videos captured by GoPro cameras showcasing furniture assembly tasks,
all recorded against a uniform background by 14 individuals.
IkeaFA categorizes these assembly actions into 12 classes.

\noindent \textbf{UAV-Human (UAV-H)~\cite{li2021uav}:}
This dataset is gathered through the lens of an Unmanned Aerial Vehicle,
offering a unique perspective through its collection of video footage.
This dataset features a variety of recording types, including fisheye and night-vision videos.
In our study, we use videos captured by standard RGB cameras.
This subset includes 22,476 videos having 155 different action categories.

\noindent \textbf{Kinetics400 (K400)~\cite{kay2017kinetics}:}
This large-scale dataset includes around 300,000 video clips,
each labeled with one of 400 actions.
The Kinetics400 videos are all sourced from YouTube and last about 10 seconds each.

\begin{table*}[t!]
\centering
\scriptsize

\caption{
\textbf{Pre-training setting for each dataset.}
}

\vspace{-2mm}

\begin{tabular}{lccc}
\toprule[1.2pt]

configuration                                   & Kinetics400 & MiniSSV2             & Other Datasets    \\
\midrule[0.5pt]
optimizer                                       & \multicolumn{3}{c}{AdamW~\cite{loshchilov2017decoupled}}                        \\
learning rate                                   & \multicolumn{3}{c}{1e-3}                                                   \\
weight decay                                    & \multicolumn{3}{c}{0.05}                                                      \\
optimizer momentum                              & \multicolumn{3}{c}{ $\beta_1=0.9, \beta_2=0.95$}                            \\
mask ratio                                     & \multicolumn{3}{c}{0.75}                                      \\
batch size                                      & \multicolumn{3}{c}{256}                                                       \\
batch size                                      & \multicolumn{3}{c}{256}                                                       \\
learning rate schedule                          & \multicolumn{3}{c}{cosine decay}                                              \\
warmup epochs                                   & \multicolumn{3}{c}{40}                                                         \\
epochs                                          & 800          & 2000                  & 2000                 \\
flip augmentation                               & \checkmark  & -                    & \checkmark         \\
\bottomrule[1.2pt]

\end{tabular}

\label{tab:pretraining_setting}
\end{table*}

\begin{table*}[t!]
\centering
\scriptsize

\caption{
\textbf{Fine-tuning setting for each dataset.}
}

\vspace{-2mm}

\begin{tabular}{lccc}
\toprule[1.2pt]

configuration                                   & Kinetics400 & MiniSSV2             & Other Datasets    \\
\midrule[0.5pt]
optimizer                                       & \multicolumn{3}{c}{AdamW~\cite{loshchilov2017decoupled}}                        \\
learning rate                                   & \multicolumn{3}{c}{1e-3}                                                  \\
weight decay                                    & \multicolumn{3}{c}{0.05}                                                      \\
optimizer momentum                              & \multicolumn{3}{c}{ $\beta_1=0.9, \beta_2=0.999$}                            \\
batch size                                      & \multicolumn{3}{c}{128}                                                       \\
learning rate schedule                          & \multicolumn{3}{c}{cosine decay}                                              \\
warmup epochs                                   & \multicolumn{3}{c}{5}                                                         \\
epochs                                          & 50          & 100                  & 100                 \\
repeated augmentation~\cite{hoffer2020augment}  & \multicolumn{3}{c}{2}                                                         \\
flip augmentation                               & \checkmark  & -                    & \checkmark         \\
RandAug~\cite{cubuk2020randaugment}             & \multicolumn{3}{c}{(9, 0.5)}                                                  \\
label smoothing~\cite{szegedy2016rethinking}    & \multicolumn{3}{c}{0.1}                                                       \\
mixup~\cite{zhang2017mixup}                     & \multicolumn{3}{c}{0.8}                                                       \\
cutmix~\cite{yun2019cutmix}                     & \multicolumn{3}{c}{1.0}                                                       \\
drop path~\cite{huang2016deep}                  & 0.1         & 0.1                  & 0.2                \\
dropout                                         & 0.0         & 0.0                  & 0.5                \\
layer-wise lr decay~\cite{bao2021beit}          & \multicolumn{3}{c}{0.75}                                                      \\
sampling                                        & dense sampling~\cite{feichtenhofer2019slowfast, wang2018non} & uniform sampling~\cite{wang2018temporal} & dense sampling \\

\bottomrule[1.2pt]

\end{tabular}

\label{tab:finetuning_setting}
\end{table*}

\section{Implementation Details}
\label{sec:implementation}

We conducted the experiments with 8 A100 GPUs for both pre-training and fine-tuning,
mostly following the settings in VideoMAE~\cite{tong2022videomae}.
The settings for pre-training are detailed in~\cref{tab:pretraining_setting}
and those for fine-tuning are described in~\cref{tab:finetuning_setting}.
We used PyTorch~\cite{paszke2019pytorch} to implement our framework.

\section{Pseudo-code of Pseudo Motion Generator (PMG)}
\label{sec:pmg}

While the algorithm of our Pseudo Motion Generator (PMG) is detailed in the main paper,
we offer Python pseudo-code for PMG in~\cref{fig:codes} for more clarity.

\section{Parameters of Image Augmentations in PMG}
\label{sec:parameter}

Since it is difficult to find the optimal parameters for each image augmentation in our framework,
we implement each augmentation with a predefined range of parameters as follows:

\begin{itemize}
    \item \textbf{Sliding Window: }
         Cut a $112\times112$ window from a $224 \times 224$ image and move it randomly.  

    \item \textbf{Zoom-in/out:}
        For Zoom-out, randomly set a window from a $224\times224$ image
        within the size range of [0.2, 0.45], then gradually enlarge the window
        until it reaches a random size between [0.55, 0.95].
        For Zoom-in, reverse the process for pseudo-motion videos generated by Zoom-out.
        We randomly choose between Zoom-in and Zoom-out with a 50\% probability.

    \item \textbf{Fade-in/out: }
        For Fade-out, make an input image gradually become completely invisible.
        For Fade-in, reverse the process of pseudo-motion videos generated by Zoom-in.
        We randomly choose between Fade-in and Fade-out with a 50\% probability.

    \item \textbf{Affine Transformation}
        We use the AffineTransformation class provided in PyTorch~\cite{paszke2019pytorch}.
        The rotation angle in degrees is randomly selected between -15 and 15.
        The translation is randomly selected between [-0.01, 0.01] for both horizontal and vertical directions.
        The scale value is randomly selected between [0.9999, 1.0001].
        The shear angle value in degrees is randomly selected between -1 and 1.

    \item \textbf{Perspective Transformation}
        We use the PerspectiveTransformation class provided in PyTorch.
        The scale of distortion is set to 0.05.
        
    \item \textbf{Color Jitter}:
        We use the ColorJitter class provided in PyTorch.
        We set the range of brightness as [0.0, 0.2], that of contrast as [0, 0.3],
        that of saturation [0, 0.2], that of hue [0.0, 0.1].

    \item \textbf{CutMix}:
        As in Sliding Window, we cut a $112\times112$ window from an image and
        paste it to another $224 \times 224$ image,
        then move the window randomly.
\end{itemize}

We understand that these predefined parameters are not optimal and there is room for further consideration.
We plan to conduct exhaustive experiments and develop a framework that does not rely on hand-crafted augmentations.


\begin{figure}[t!]
\centering

\begin{lstlisting}[language=Python]
import random

transform_list = [
    "Identity", "Sliding Window", "Zoom-in", "Zoom-out",
    "Fade-in", "Fade-out", "Affine Transformation",
    "Perspective Transformation", "Color Jitter", "CutMix",
]

def generate_pseudo_motion(image, T):
    """Pseudo Video Generator.

    Args:
        image: Input image.
        T: The number of frames in a video.
    """
    transform = random.choice(transform_list)
    params = transform.get_random_parameters()

    video = [image]
    previous_frame = image
    for _ in range(T - 1):
        transformed_frame = transform(previous_frame, params)
        video.append(transformed_frame)
        previous_frame = transformed_frame

    return video
\end{lstlisting}

\vspace{-3mm}
\caption{\textbf{Python pseudo-code for Pseudo Motion Generator (PMG).}}
\label{fig:codes}

\end{figure}

\begin{figure}[t]
\centering
\footnotesize
\scriptsize

\begin{subfigure}[b]{0.9\textwidth}
\centering
\includegraphics[width=\textwidth]{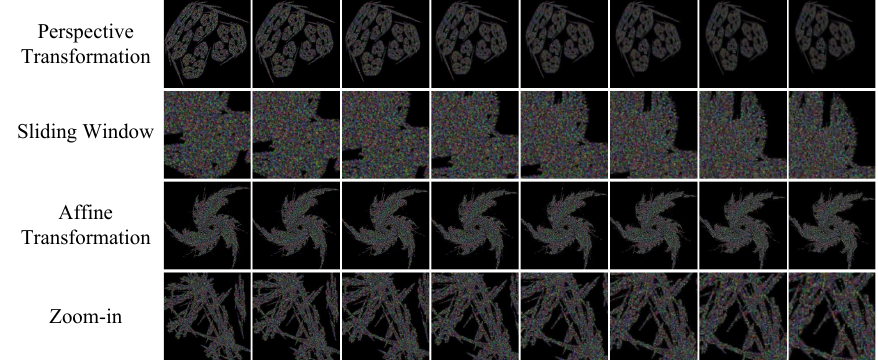}
\caption{FractalDB}
\label{fig:fractal_exapmle}
\end{subfigure}

\vspace{5mm}

\begin{subfigure}[b]{0.9\textwidth}
\centering
\includegraphics[width=\textwidth]{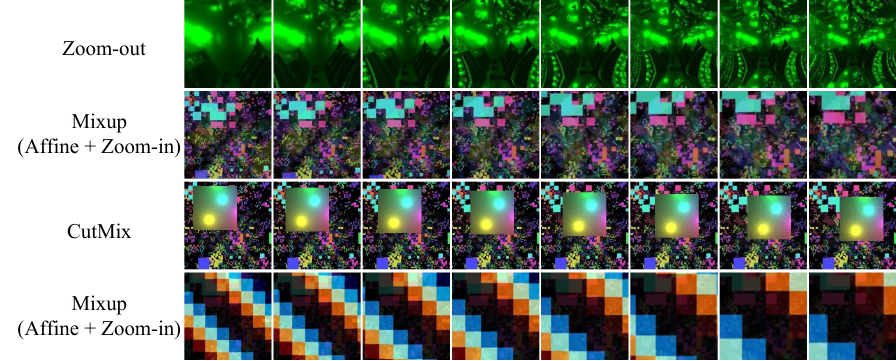}
\caption{Shaders1k}
\label{fig:shaders1k_example}
\end{subfigure}

\vspace{5mm}

\begin{subfigure}[b]{0.9\textwidth}
\centering
\includegraphics[width=\textwidth]{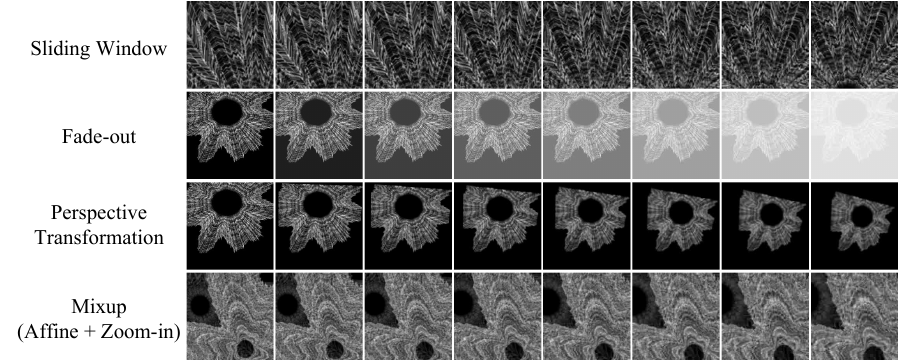}
\caption{VisualAtom}
\label{fig:visual_atom_example}
\end{subfigure}
\vspace{-1mm}

\caption{\textbf{Examples of pseudo-motion videos generated from synthetic image datasets.}}
\label{fig:supp_example}
\end{figure}

\section{Examples of Pseudo-motion Videos}
\label{sec:example}

\cref{fig:supp_example} shows the examples of pseudo-motion videos
generated from three synthetic image datasets;
FractalDB~\cite{kataoka2020pre},
Shaders1k~\cite{baradad2022procedural},
and Visual Atom~\cite{takashima2023visual}.
Although the appearance and motions in these videos differ from real videos,
they exhibit a wide range of motion and appearance patterns.
This variety enables VideoMAE to learn effectively.
Specifically, pre-training with pseudo-motion videos generated from Shaders1k
improves the model's performance compared to pre-training with those from the other sources.
This improvement is attributed to the videos from Shaders1k having a clear correspondence
of patches between frames, which suits for VideoMAE.

\begin{figure}[t]
    \centering
    \footnotesize
    \includegraphics[height=0.94\textheight]{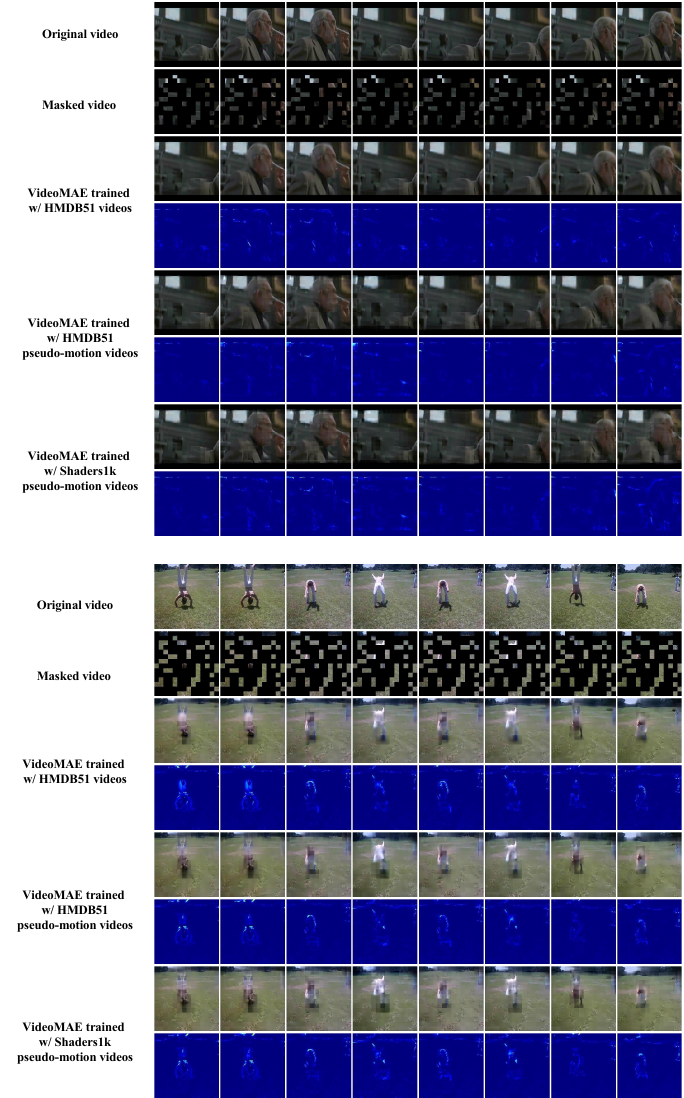}

    \caption{
        \textbf{Visualization of outputs and loss heatmaps for VideoMAE on HMDB51.}
        The mask ratio is set as 75\%.
        Loss heatmaps are normalized per frame.
    }
    \label{fig:hmdb_visualization}

\end{figure}

\begin{figure}[t]
    \centering
    \footnotesize
    \includegraphics[height=0.94\textheight]{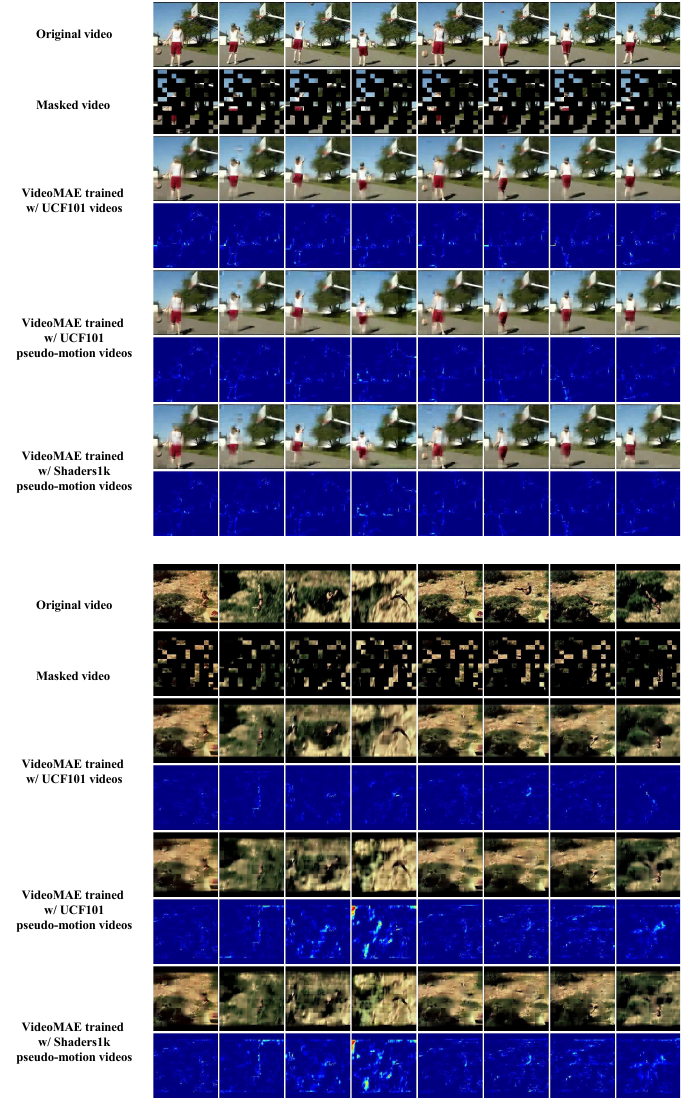}

    \caption{
        \textbf{Visualization of outputs and loss heatmaps for VideoMAE on UCF101.}
        The mask ratio is set as 75\%.
        Loss heatmaps are normalized per frame.
    }
    \label{fig:ucf_visualization}

\end{figure}


\section{Quantitative Results of Our Framework}
\label{sec:quantitative}

To verify that VideoMAE successfully learns the reconstruction task,
we visualized its output results on HMDB51 and UCF101.
We compared the outputs of three models:
(i) VideoMAE trained on real videos from each dataset,
(ii) VideoMAE trained on pseudo-motion videos generated from frames on each video dataset,
and (iii) VideoMAE trained on pseudo-motion videos from Shaders1k.
\cref{fig:hmdb_visualization} and~\cref{fig:ucf_visualization}
shows the results for HMDB51 and UCF101, respectively.
The inputs for these models were sampled from the test set,
which was not used for pre-training.
Despite not being trained on real videos,
VideoMAE trained on Shaders1k manages to achieve a reasonable level of accuracy
in reconstructing real videos.
This suggests that the method can roughly capture
the complex motion and shape characteristics of the real world.

However, compared to VideoMAE trained on real videos,
VideoMAE trained on pseudo-motion videos
struggles with the reconstruction of finer details.
This issue likely arises because our PMG applies image transformations globally,
hindering its ability to learn fine-grained motions.
Consequently, our framework exhibits lower performance
in classifying certain fine-grained actions,
compared to VideoMAE trained on real videos
(See~\cref{sec:failure_cases}).

\section{Failure Cases}
\label{sec:failure_cases}

\begin{figure}
    \centering
    \includegraphics[height=0.95\textheight]{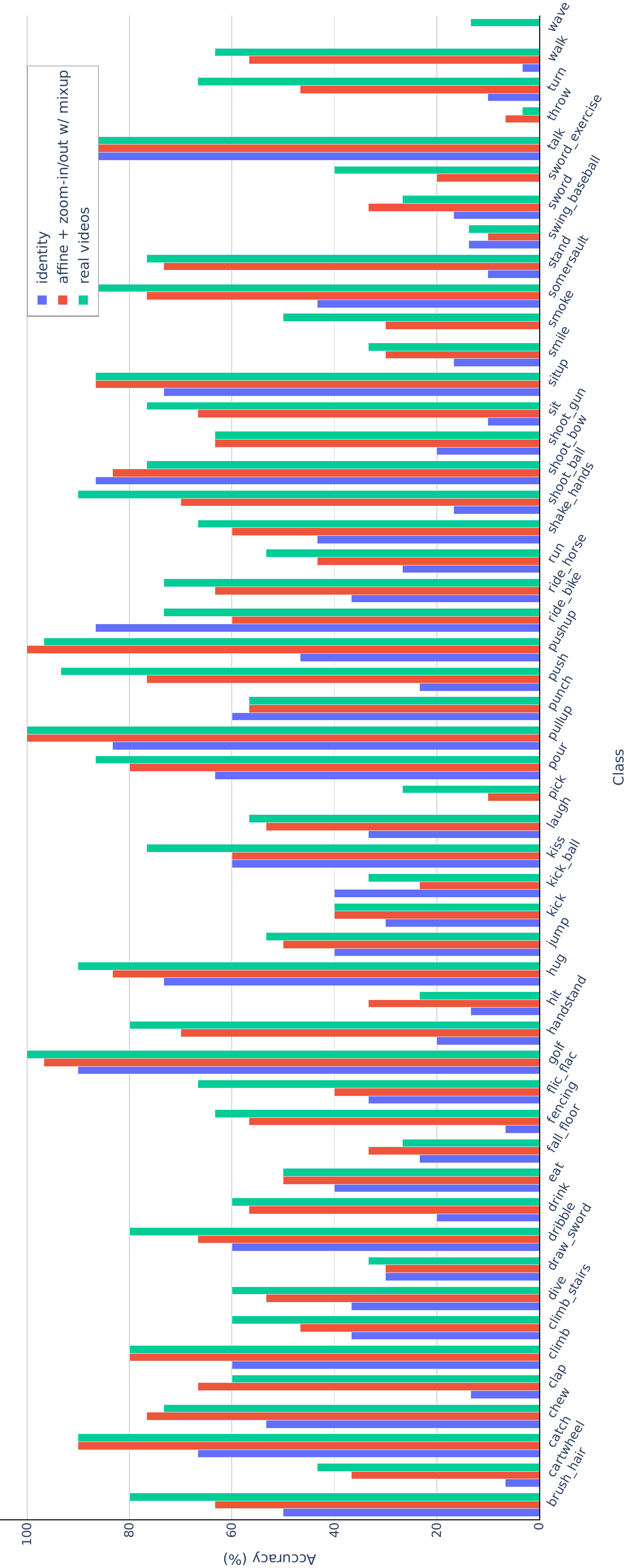}

    \caption{\textbf{Comparison of accuracy per class for each model on HMDB51.}}
    \label{fig:acc_per_class_hmdb51}
\end{figure}


\begin{figure}[t]
\centering
\footnotesize
\begin{subfigure}[b]{0.48\textwidth}
\centering
\includegraphics[width=\textwidth]{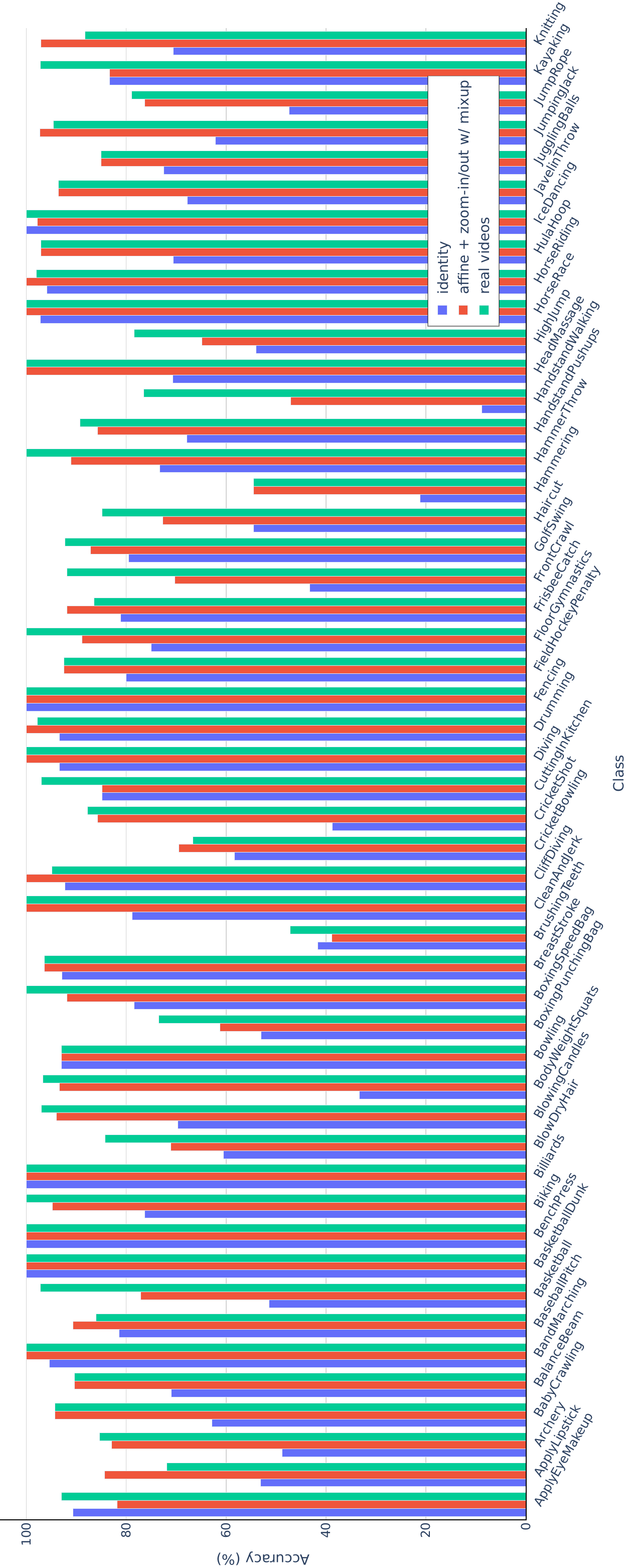}
\label{fig:epoch_vs_acc}
\end{subfigure}
\hfill
\begin{subfigure}[b]{0.48\textwidth}
\centering
\includegraphics[width=\textwidth]{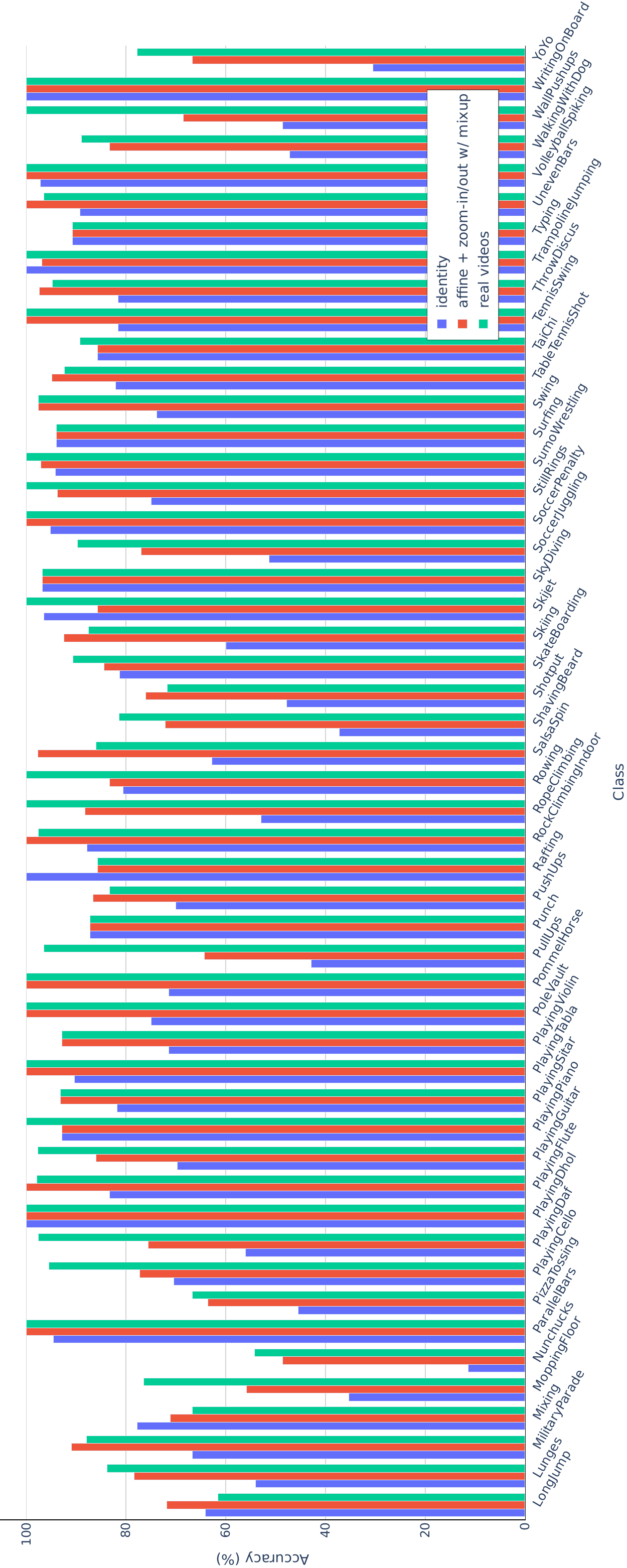}
\label{fig:class_vs_acc}
\end{subfigure}
\caption{\textbf{Comparison of accuracy per class for each model on UCF101.}}
\label{fig:acc_per_class_ucf101}
\end{figure}

We further analyzed the failure cases of our framework
compared to VideoMAE when trained with real videos.
For this analysis, we evaluated three models:
(i) VideoMAE trained with pseudo-motion videos by Identity (no-motion videos),
(ii) VideoMAE trained with pseudo-motion videos by Affine Transformation and Zoom-in/out combined with Mixup.
(iii) VideoMAE trained with real videos.

\cref{fig:acc_per_class_hmdb51} presents the accuracy per class on HMDB51.
Between model (i) and (ii),
model (ii) demonstrated improved performance of actions such as 'cartwheel', 'sit', and 'stand',
which rely on motion information for recognition.
However, in the comparison between model (ii) and (iii),
we found that model (ii) struggled to classify actions like 'kiss', 'push', 'shake hands', and 'wave',
which involve more subtle and fine-grained motion.

\cref{fig:acc_per_class_ucf101} shows the accuracy per class on UCF101.
As in the patterns observed in HMDB51,
model (ii) improved the performance in classes like
'BodyWeightSquats', 'CleanAndJerk', 'JumpRope' and 'YoYo',
where videos lack object and background cues.
Additionally, model (ii) successfully differentiated between action classes involving similar objects,
for instance, 'BasketballDunk' versus 'Basketball', and 'HammerThrow' versus 'Hammering'.
However, in comparison between model (ii) and (iii),
we found it was difficult for model (ii) to recognize more fine-grained actions such as
'Handstand Walking', 'Nunchucks', 'PullUps', and 'WallPushups'.

Our framework struggles to capture fine-grained motion information.
since our PMG applies hand-crafted image transformations globally.
Consequently, the model trained by our framework has difficulty recognizing
fine-grained actions, representing one of the limitations of our framework.
Addressing this issue will be a priority for our future work.

\section{Linear Probing}
\label{sec:linear_probing}

\begin{table}[t]

\centering
\scriptsize

\caption{
    \textbf{Results on SynAPT benchmark in the linear probing setting.}
    $^\dag$ Results reported in~\cite{kim2022transferable}.
}
\label{tab:synapt_linear}
\vspace{-1mm}
\begin{tabular}{cccc|cccccc}
\toprule[1.2pt]
\multirow{2}{*}{Method}       & \multicolumn{3}{c|}{Pre-training}                                         & \multicolumn{6}{c}{Downstream Tasks}                   \\
                              & Dataset                                                 & \#data     & ~labels    & UCF101 & HMDB51 & MiniSSV2 & Diving48 & IkeaFA & UAV-H \\
\midrule[0.5pt]
TimeSformer$^\dag$            & \begin{tabular}{c}IN-21k\\+Synthetic\end{tabular}       & 150k       & \checkmark & 82.1   & 49.2   & 21.2      & 19.2    & 45.5   & 13.8      \\
\midrule[0.5pt]
PPMA~\cite{zhong2023learning} & \begin{tabular}{c}NH-Kinetics\\+Synthetic\end{tabular}  & 300k       & \checkmark & 88.4   & 64.9   & 34.9      & 21.9    & 57.7   & 19.3      \\
\midrule[0.5pt]
Ours                 & Shaders1k                                                        & 100k       &            & 42.5   & 28.0   & 10.3      & 6.4     & 33.1   & 1.1       \\
\bottomrule[1.2pt]

\vspace{-3mm}
\end{tabular}
\end{table}


Another limitation of our framework is that
our framework does not learn high-level semantic features,
because our framework focuses on low-level features and
does not utilize labels during pre-training, unlike PPMA~\cite{zhong2023learning}.
This limitation leads to lower performance in the linear probing settings,
where the weights of the encoder are frozen while only the linear layer is trained (\cref{tab:synapt_linear}).
Moreover, it is challenging to extend our framework
to other tasks like video-text retrieval and video captioning,
without additional training or extra labeled data.
We will also tackle this issue in future work.

\end{document}